\documentclass[lettersize,journal]{IEEEtran}
\usepackage{amsmath,amsfonts}
\usepackage{algorithmic}
\usepackage{array}
\usepackage[caption=false,font=normalsize,labelfont=sf,textfont=sf]{subfig}
\usepackage{textcomp}
\usepackage{stfloats}
\usepackage{url}
\usepackage{verbatim}
\usepackage{graphicx}
\usepackage{soul}
\usepackage{url}
\usepackage[hidelinks]{hyperref}
\usepackage[utf8]{inputenc}
\usepackage[small]{caption}
\usepackage{booktabs}
\usepackage{times}
\usepackage{cite}
\usepackage{algorithm}
\usepackage{color, xcolor}
\usepackage{multirow}
\usepackage{colortbl}
\usepackage{amssymb}
\usepackage{pifont}
\newcommand{\cmark}{\ding{51}}%
\newcommand{\xmark}{\ding{55}}%
\usepackage{hyperref}
\hypersetup{
    colorlinks=true,
    linkcolor=blue,
    filecolor=blue,      
    urlcolor=blue,
    citecolor=blue,
}

\usepackage[normalem]{ulem}

\usepackage{caption}
\def\BibTeX{{\rm B\kern-.05em{\sc i\kern-.025em b}\kern-.08em
    T\kern-.1667em\lower.7ex\hbox{E}\kern-.125emX}}
\usepackage{balance}

\begin{document}
\title{CarDD: A New Dataset for Vision-based Car Damage Detection}

\author{Xinkuang Wang, Wenjing Li, Zhongcheng Wu
\thanks{
Manuscript created *; revised *; accepted *.
This work was supported by the Hefei Comprehensive National Science Center through the Pre-research Project on Key Technologies of Integrated Experimental Facilities of Steady High Magnetic Field and Optical Spectroscopy and the High Magnetic Field Laboratory of Anhui Province.
The Associate Editor for this article was ***.
(\textit{Corresponding Author: Wenjing Li}.)

Xinkuang Wang, Wenjing Li, and Zhongcheng Wu are with High Magnetic Field Laboratory, HFIPS, Chinese Academy of Sciences, Hefei 230031, China, and the University of Science and Technology of China, Hefei 230026, China. They are also with High Magnetic Field Laboratory of Anhui Province, Hefei 230031, China. (e-mail: wangxk0624@mail.ustc.edu.cn; wjli007@mail.ustc.edu.cn).}}

\markboth{Journal of \LaTeX\ Class Files,~Vol.~18, No.~9, March~2022}%
{How to Use the IEEEtran \LaTeX \ Templates}

\maketitle

\begin{abstract}
Automatic car damage detection has attracted significant attention in the car insurance business. However, due to the lack of high-quality and publicly available datasets, we can hardly learn a feasible model for car damage detection.
To this end, we contribute with Car Damage Detection (CarDD), the first public large-scale dataset designed for vision-based car damage detection and segmentation. Our CarDD contains 4,000 high-resolution car damage images with over 9,000 well-annotated instances of six damage categories (examples are shown in Figure~\ref{fig:Visualization}). We detail the image collection, selection, and annotation processes, and present a statistical dataset analysis. Furthermore, we conduct extensive experiments on CarDD with state-of-the-art deep methods for different tasks and provide comprehensive analyses to highlight the specialty of car damage detection.
CarDD dataset and the source code are available at \textit{\url{https://cardd-ustc.github.io}}.
\end{abstract}


\section{Introduction}
\label{sec:intro}
\IEEEPARstart{A}{utomatic} car damage assessment systems can effectively reduce human efforts and largely increase damage inspection efficiency.
Deep-learning-aided car damage assessment has thrived in the car insurance business in recent years \cite{01}.
Regular claims involving minor exterior car damages (\textit{e.g.}, scratches, dents, and cracks) can be detected automatically, sparing a tremendous amount of time and money for car insurance companies. Therefore, the research on car damage assessment gains many concerns.

Vision-based car damage assessment aims to locate and classify the damages on the vehicle and visualize them by contouring their exact locations. 
Car damage assessment is closely linked to the task of visual detection and segmentation, where deep neural networks have shown a promising capability \cite{24,19}.
Several studies have attempted to adopt neural networks to tackle the tasks of car damage classification \cite{02,03,04,05,06}, detection \cite{07,08,09}, and segmentation \cite{10,11}.
For the classification problem, some previous work \cite{02,03,06} applies CNN-based models to identify the damage category. 
And for the object detection and instance segmentation task, YOLO \cite{24}, and Mask R-CNN \cite{19} are mostly referenced, respectively. 

However, due to the insufficient datasets for training such models, the research on vehicle damage assessment is far behind.
Although the dataset presented by \cite{14} 
is publicly available, it can only be used for damage type classification.
Recall that, the objective of car damage assessment is damage detection and segmentation.
Existing works are often conducted on private datasets \cite{02,03,05,06,07,08,09,10,11}, and hence the performance of different approaches can not be fairly compared, further hindering the development of this field.

Therefore, to facilitate the research of vehicle damage assessment, in this paper, we introduce a novel dataset for \textbf{Car} \textbf{D}amage \textbf{D}etection, namely \textbf{CarDD}.
CarDD offers a variety of potential challenges in car damage detection and segmentation
and is the first publicly available dataset, which combines the following properties:

\begin{itemize}
   \item \textbf{Damage types:} \textit{dent, scratch, crack, glass shatter, tire flat,} and \textit{lamp broken}.
    \item \textbf{Data volume:} it contains 4,000 car damage images with over 9,000 labeled instances
    which is the largest public dataset of its kind to our best knowledge.
     \item \textbf{High-resolution:} 
     the image quality in CarDD is far greater than that of existing datasets. The average resolution of CarDD is 684,231 pixels (13.6 times higher) compared to 50,334 pixels of the GitHub dataset \cite{14}.
    \item \textbf{Multiple tasks:} our dataset comprises four tasks: classification, object detection, instance segmentation, and salient object detection (SOD).
    \item \textbf{Fine-grained:} different from general-purpose object detection and segmentation tasks, the difference between car damage types like \textit{dent} and \textit{scratch} is subtle.
    \item \textbf{Diversity:} images of our dataset are more diverse regarding object scales and shapes than the general objects due to the nature of the car damage.
\end{itemize}

In summary, we make three major contributions in this paper:
\begin{itemize}
    \item First, we release the largest  
    dataset called CarDD, the first publicly available dataset for car damage detection and segmentation. CarDD contains high-quality images annotated with damage type, damage location, and damage magnitude, which is more practical than existing datasets. CarDD helps train and evaluate algorithms for car damage detection, offering new opportunities to advance the development of car damage assessment.
    \item Second, we conduct extensive exploratory experiments on object detection and instance segmentation tasks and carefully analyze the results, which offers valuable insights to the researchers.
    The unsatisfactory results of state-of-the-art (SOTA) methods on CarDD reveal the difficulty of the car damage assessment problem, proposing new challenges to the tasks of object detection and instance segmentation.
    To address these challenges, we introduce an improved method called DCN+, which significantly boosts the detection and segmentation performance and outperforms all SOTA methods. 
    \item Third, we are the first to leverage SOD methods to deal with this task, and the experimental results reveal that SOD methods have the potential to solve the car damage assessment, especially on the irregular damage types.
    We hope CarDD can benefit the vision-based car damage assessment and contribute to the computer vision community.
\end{itemize}

\begin{figure*}
  \centering
  \hsize=\textwidth
   \includegraphics[width=1\textwidth, height=0.55\textwidth]{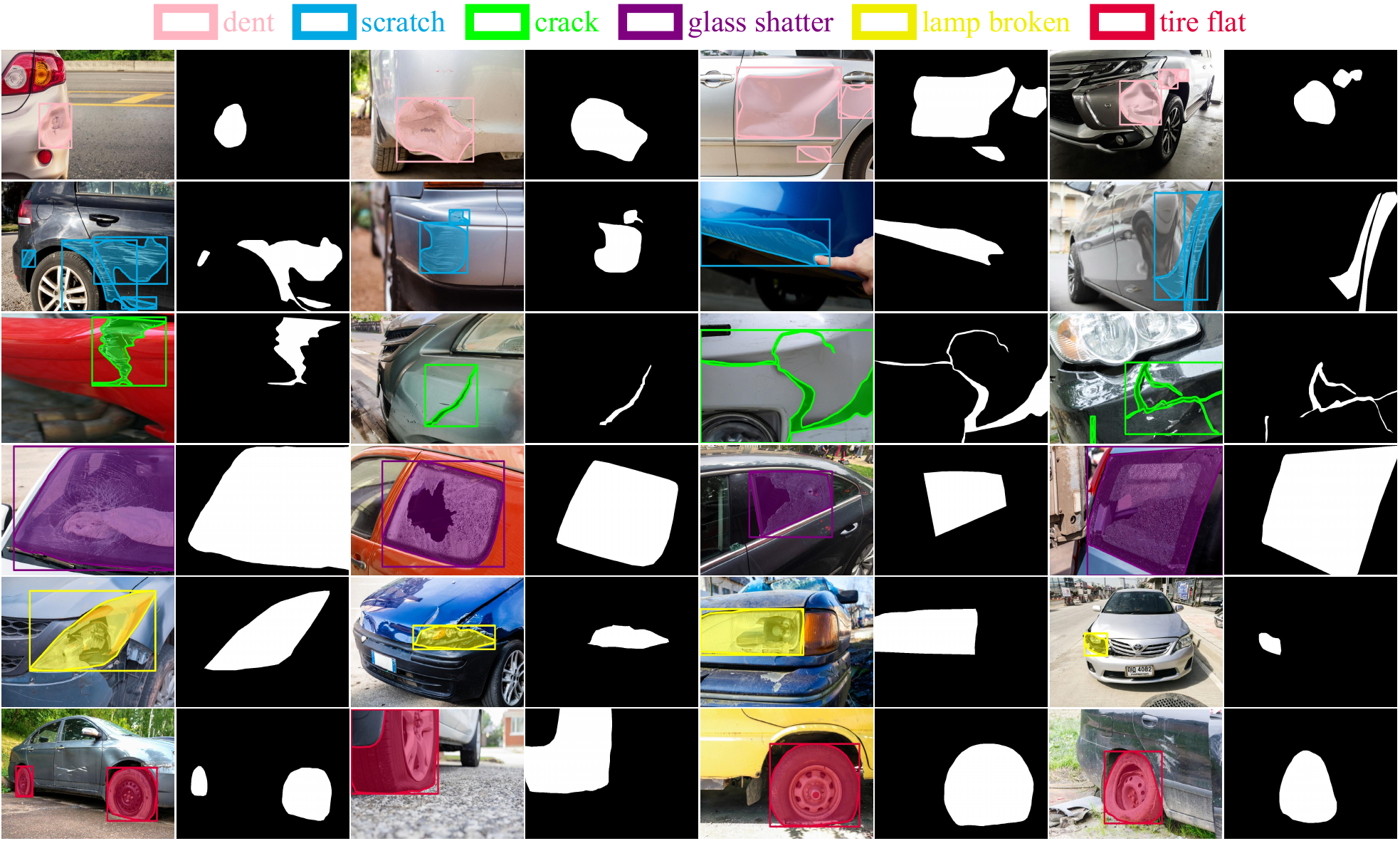}
   \caption{Samples of annotated images in the CarDD dataset. Note that masks of only a single type of damage are activated in each image for a clear demonstration. Each instance is assigned a unique ID in the annotation files.}
   \label{fig:Visualization}
   \vspace{-4mm}
\end{figure*}

\section{Related Work}
\label{sec:related-work}

Deep learning has immensely pushed forward studies involving the intact (undamaged) car, such as fine-grained vehicle recognition and retrieval \cite{fine-grained, vehicle-instance-retrieval}, vehicle type and component classification \cite{vehicle-type, taillight}, license plate recognition \cite{licence-plate-1, licence-plate-2}.
However, the detection of damaged components is a more complicated problem for the following reasons.
Firstly, the damaged car has a different appearance from the original one, which means the models trained on the dataset of undamaged vehicles are no longer effective.
Secondly, damages are quite different from the daily seen objects, bringing new challenges to existing algorithms in the general object detection field.

\subsection{Car Damage Related Methods}
Early works are based on heuristics.
For instance, Jayawardena \cite{12} compared the ground-truth undamaged 3D CAD car model with the damaged car photographs to detect mild damages. 
Yet, the performance of this method will dramatically degrade in a realistic application scenario
since the appearance of the car is ever-changing.
Gontscharov \textit{et al.} \cite{13} installed the developed sensors on the vehicle that can detect structural damage by collecting structure-borne sound. 
However, this method is not practical due to the high-cost sensors, as well as the complexity of installation.
Thus, while these methods give a good try to solve the problem of automatic car damage detection, they have not been widely applied.

The success of deep learning spread to the car damage field in 2017, when Patil \textit{et al.} \cite{06} employed a CNN model to classify car damage types (\textit{e.g.}, bumper dent, door dent).
After that, \cite{02} and \cite{03} also use CNN-based models to classify the image into a limited number of car damage types. 
Nevertheless, simple damage classification can not satisfy the actual needs for two reasons.
First, damage classification is hard to handle a sample with multiple types of damage.
Second, damage classification can not tell the damage location and damage magnitude.
Therefore, work studying car damage detection and segmentation emerges. 
Dwivedi \textit{et al.} \cite{07} applied YOLOv3 to detect the damaged region. 
Singh \textit{et al.} \cite{11} experimented with Mask R-CNN and PANet and a transferred VGG16 network to localize and detect various classes of car components and damages.
However, the scores such as mAP in their work are relatively low, indicating the difficulty of the car damage detection task.

Besides the research community, some commercial applications also tried to recognize damaged components and the damage degree from images\footnote{\url{https://tractable.ai/en/industries/auto-claims}} \footnote{\url{https://skyl.ai/models/vehicle-damage-assessment-using-ai}} \footnote{\url{https://www.altoros.com/solutions/car-damage-recognition}}.
Recently, the Ant Financial Services Group \cite{01} 
proposed to adopt object detection and instance segmentation techniques combined with frame selection algorithms to detect damaged components and segment accurate component boundaries.
These applications prove the importance and urgent demand for car damage detection techniques in the car insurance industry.

\begin{table*}[!t]
\normalsize
  \centering
 \caption{ Comparison of datasets for car damage classification, detection and segmentation.  \textbf{C:} damage classification, \textbf{D:} damage detection, \textbf{S:} damage segmentation, \textbf{SOD:} salient object detection. \textbf{PA:} Publicly Available. N/A: information not provided by the authors.}
 \setlength{\tabcolsep}{3.8pt}
  \begin{tabular}{c|c|c|c|c|c}
    \toprule
    \textbf{Dataset} & \textbf{Task} & \textbf{Size} & \textbf{PA} & \textbf{\# Categories} & \textbf{Damage Categories} \\
\hline
    Deijn \cite{02} & C & 1,007 & \xmark & 4 & \parbox[c]{8.5cm}{dent, glass, hail, and scratch} \\ 
    \hline
    Balci \textit{et al.} \cite{03} & C & 533 & \xmark & 2 & \parbox[c]{8.5cm}{damaged and non-damaged} \\ \hline
    Waqas \textit{et al.} \cite{05} & C & 600 & \xmark & 3 & \parbox[c]{8.5cm}{medium damage, huge damage, and no damage} \\ \hline
    Patil \textit{et al.} \cite{06} & C & 1,503 & \xmark & 7 & \parbox[c]{8.5cm}{bumper dent, door dent, glass shatter, head lamp broken, tail lamp broken, scratch, and smash} \\ \hline
    Neo \cite{14} & C & 1,150 &\cmark & 8 & \parbox[c]{8.5cm}{crashed, bumper dent, glass shatter, scratch, door dent, front damage, headlight damage, and tail-lamp damage} \\ \hline
    Dwivedi \textit{et al.} \cite{07} & C, D & 1,077 & \xmark & 7 & \parbox[c]{8.5cm}{smashed, scratch, bumper dent, door dent, head lamp broken, glass shatter, and tail lamp broken} \\ \hline
    Li \textit{et al.} \cite{09} & D & 1,790 & \xmark & 3 & \parbox[c]{8.5cm}{scratch, dent, and crack} \\ \hline
    Patel \textit{et al.} \cite{08} & D & 326 & \xmark &  3 & \parbox[c]{8.5cm}{bump, dent, and scratch}  \\ \hline
    Dhieb \textit{et al.} \cite{10} & D, S & N/A  & \xmark & 3 & \parbox[c]{8.5cm}{minor, moderate, and major} \\ \hline
    Singh \textit{et al.} \cite{11} & S & 2,822 & \xmark & 5 & \parbox[c]{8.5cm}{scratch, major dent, minor dent, cracked, and missing} \\
    \hline
    \textbf{CarDD}  & \textbf{C, D, S, SOD} & \textbf{4,000} & \cmark & 6 & \parbox[c]{8.5cm}{dent, scratch, crack, glass shatter, lamp broken, and tire flat} \\
    \bottomrule
  \end{tabular}

  \label{tab:datasets-comparison}
\end{table*}

\subsection{Car Damage Related Datasets}

Table~\ref{tab:datasets-comparison} summarizes the existing datasets for car damage assessment. 
The type of task, damage categories, dataset size, and availability are listed.
The tasks include car damage classification, object detection, instance segmentation, and salient object detection, abbreviated as C, D, S, and SOD, respectively. 
From Table~\ref{tab:datasets-comparison}, we obtain three observations.
First, many of these datasets are relatively small in size (under 2K) except the dataset
\cite{11} that includes 2,822 images.
We must admit that the advantage of deep learning cannot be tested to the full
since these datasets are relatively small.
Second, half of the existing datasets focus on the damage classification task.
As we aforementioned, damage classification is not practical enough, especially when it comes to works \cite{03,05} classifying damaged or not.
Third, more importantly, apart from \cite{14}, most of these datasets are not released,
which may hinder other researchers from following their work, and the performances of different methods can not be fairly compared.

Although 
the GitHub repository \cite{14} reveals a car damage dataset,
they did not offer further labels of the damage types and damage locations, preventing the potential of damage detection and instance segmentation.
Besides, images in this dataset are relatively low-resolution, which largely restricts the annotation quality as demonstrated in Section~\ref{sec:dataset-statistics}.

In comparison to these datasets, the CarDD dataset includes 4,000 well-annotated high-resolution samples with over 9,000 labeled instances for damage classification, detection, segmentation, and salient object detection (SOD)\cite{sod}. 
For the object detection and instance segmentation tasks, we annotate our dataset with masks and bounding boxes assigned with damage types, including \textit{dent, scratch, crack, glass shatter, lamp broken}, and \textit{tire flat}. And for the SOD task, we offer the pixel-level ground truths containing highlighted salient objects.
We release our CarDD dataset with unique characteristics (large data volume, high-quality images, fine-grained annotations) for academic research, aiming to push the field of car damage assessment
further, and opening new challenges for general object detection, instance segmentation, and salient object detection.

\section{The CarDD Dataset}
\label{sec:dataset}

We aim to provide a large-scale high-resolution car damage dataset with comprehensive annotations that can expose the challenges of car damage detection and segmentation.
This section will present an overview of the image collection, selection, and annotation process, followed by the analysis of dataset characteristics and statistics.

\begin{figure*}[t]
  \centering
   \includegraphics[width=1\textwidth]{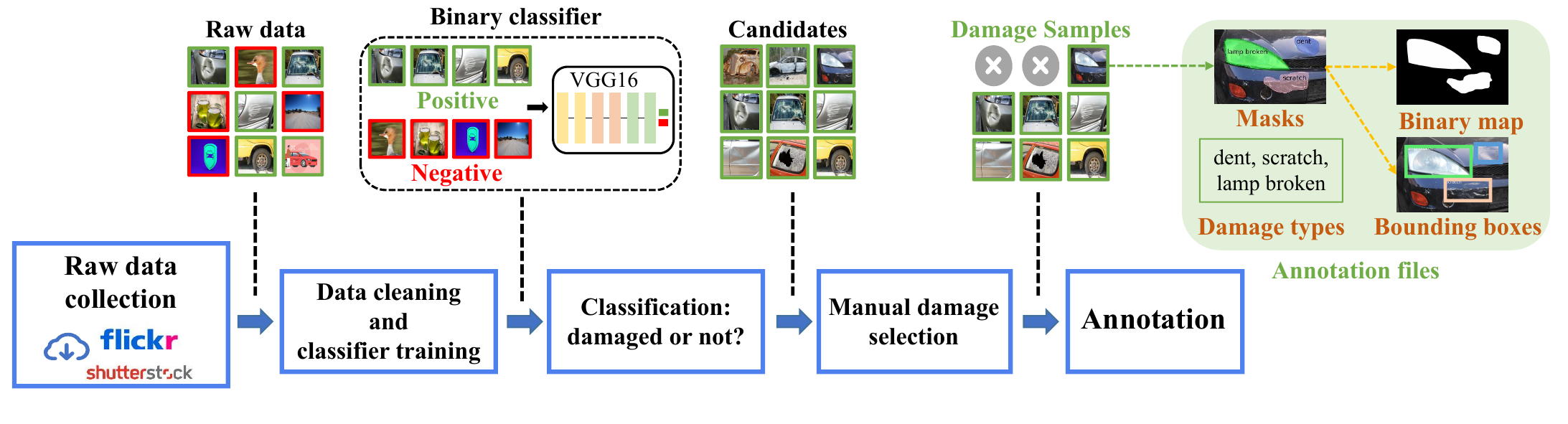}
   \caption{The generation pipeline of CarDD includes raw data collection, candidate selection, manual check, and annotation. In the candidate selection step, we train a binary classifier to determine whether an image contains damaged cars.}
   \label{fig:Pipeline}
   \vspace{-6mm}
\end{figure*}

\subsection{Dataset Construction}
\label{sec:image-collection}

\noindent {\bf Damage Type Coverage.} 
CarDD mainly focuses on external damages on the car surface that have the potential to be detected and assessed by computer vision technology.
Frequency of occurrence and clear definition are the two critical considerations for the coverage of CarDD.
Based on the statistics of Ping An Insurance Company\footnote{One of the largest insurance companies in China.}, six classes of most frequent external damage types (\textit{dent, scratch, crack, glass shatter, tire flat,} and \textit{lamp broken}) are covered in CarDD. 
These classes have relatively clear definitions compared to ``smash,'' a mixture of damages.

\noindent {\bf Raw Data Collection.} 
The dataset generation pipeline is 
illustrated in Figure~\ref{fig:Pipeline}.
To obtain high-quality raw data,
we crawled a vast number of images from Flickr\footnote{\url{https://www.flickr.com}} and Shutterstock\footnote{\url{https://www.shutterstock.com}}. 
One characteristic of those 
websites is that they provide high-quality
images and rich diversity of damage types and viewpoints.
This leads the raw image dataset to cover
a wide range of variations in terms of 
light conditions, camera distance, and shooting angles.

We picked out duplicated images automatically using the powerful tool Duplicate Cleaner \cite{17} and then manually double-checked before deleting them.
For efficiency purposes, we use the first batch of manually-selected car damage images (500 images) to train a binary classifier based on VGG16 \cite{26} to determine whether an image contains damaged cars.
The classifier classifies newly collected images as positive or negative, and the positive samples become the candidate images.

With the help of the binary classifier, we further collected over 10,000 candidate images, among which some of the categories are not in the scope of our work, for instance, the rusty, burned, or smashed cars.
We decide to attach these images separately to the dataset without annotations in case of further research. 
We manually pick out the six classes of damages from the candidates, and finally, 4,000 damage samples are passed to the annotation step.

\noindent {\bf Copyright and Privacy.}
Like many popular image datasets (\textit{e.g.,} ImageNet \cite{ImageNet}, COCO \cite{25}, LFW \cite{LFW}), CarDD does not own the copyright of the images. 
The copyright of images belongs to Flickr and Shutterstock. A researcher may get access to the dataset provided that they first agree to be bound by the terms in the licenses of Flickr and Shutterstock.
A researcher accepts full responsibility for using CarDD and only for non-commercial research and educational purposes.
Additionally, taking user privacy into account, the pictures containing user information (human faces, license plates) are mosaicked or directly deleted.

\subsection{Image Annotation}
\label{sec:annotation}

\begin{figure}[!t]
  \centering
    \includegraphics[width=\linewidth]{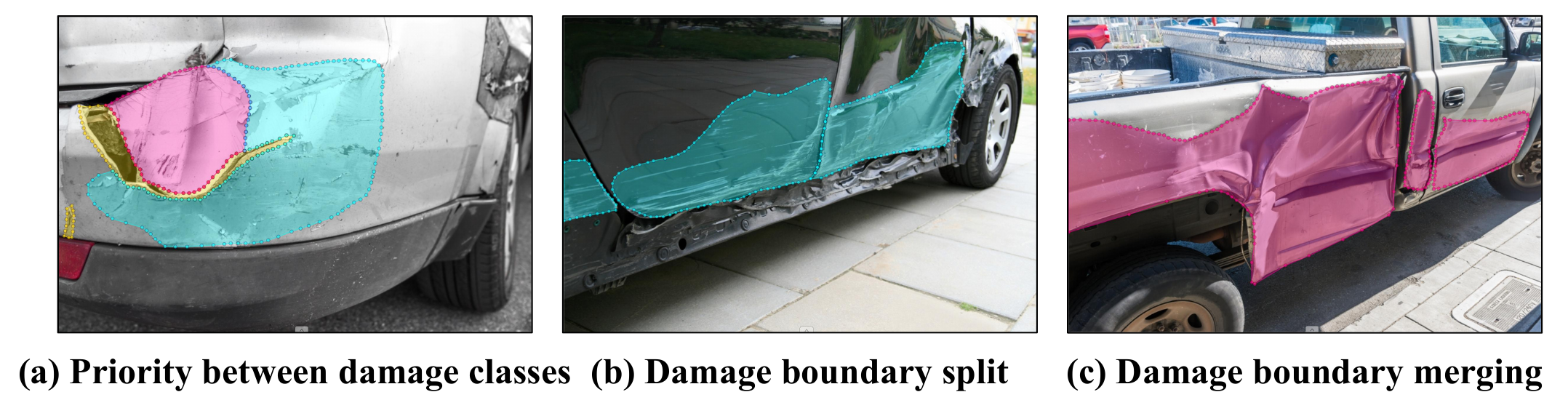}
  \caption{Examples of annotation guidelines: (a) Priority between damage classes: \textit{crack\textgreater  dent\textgreater  scratch}, (b) damages across different car components are annotated as several instances along the component edges, and (c) the same class of adjacent damages on the same component are annotated as one instance.}
  \label{fig:Guidelines}
  \vspace{-6mm}
\end{figure}

\noindent {\bf Annotation Guidelines.} 
The problems of multiplicity and ambiguity are the two main challenges in annotating car damage images.
So we formulated the annotation guidelines
according to the insurance claim standards provided by Ping An Insurance Company.
We list the three most important rules in the annotation process of CarDD:

\begin{itemize}
   \item \textbf{Priority between damage classes:} Mixed damages are very common forms on damaged cars, especially among dent, scratch, and crack. The rule to handle these images is to annotate by specified priority. For instance, in Figure~\ref{fig:Guidelines}(a), a dent, a scratch, and a crack are intertwined, and we annotate this image by the priority of \textit{crack\textgreater  dent\textgreater  scratch}. This logic is borrowed from the existing car damage repair standards, \textit{i.e.,} the crack usually requires the operation of welding, and the cracked surface with the missing part needs to be replaced, which makes the crack the severest damage of these three types. 
    The scratch requires paint repair, while the dent usually requires both metal restoration and paint repair operations, which gives a higher priority to the dent than the scratch.
    \item \textbf{Damage boundary split:} According to the car insurance claim rules formulated by Ping An Insurance Company, every damaged component is charged separately. For instance, if the front door and back door are both scratched (as shown in Figure~\ref{fig:Guidelines}(b) ), the repair department will charge for the two doors respectively. Therefore in our annotation, damages across different car components are annotated as several instances along the component edges, as shown in Figure~\ref{fig:Guidelines}(b).
    \item \textbf{Damage boundary merging:} In the car insurance claim, the same class of damages on the same component is charged at the same price regardless of the amount of the damages. Therefore, the same class of adjacent damages on the same component are annotated as one instance. For instance, although the dents on the truck carriage have clear boundaries along the wrinkles on the metal surface, we annotate all the dents on the carriage as one instance (as shown in  Figure~\ref{fig:Guidelines}(c) ).
\end{itemize}

\begin{figure*}[t]
  \centering
  \includegraphics[width=1\textwidth, height= 0.22\textwidth]{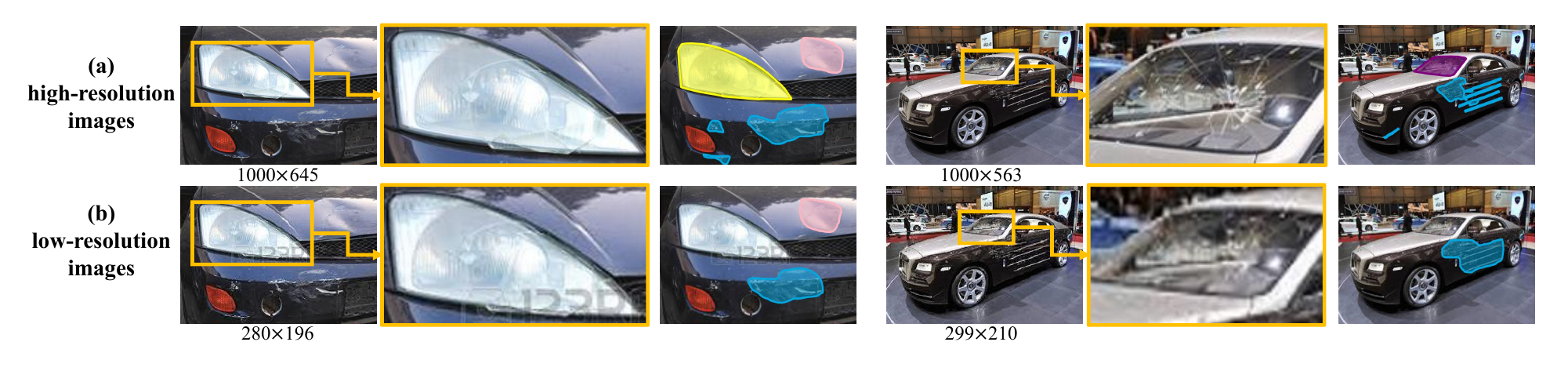}
  \caption{The effect of image resolution on the annotation quality. The first row shows high-resolution images, and the images in the second row are low-resolution images. 
  Annotations on high-resolution images can provide more damaged objects because more object-related legible textures and contours are involved in such images.}
  \label{fig:Resolution}
  \vspace{-6mm}
\end{figure*}

\begin{figure*}[!t]
  \centering
   \includegraphics[width=\textwidth]{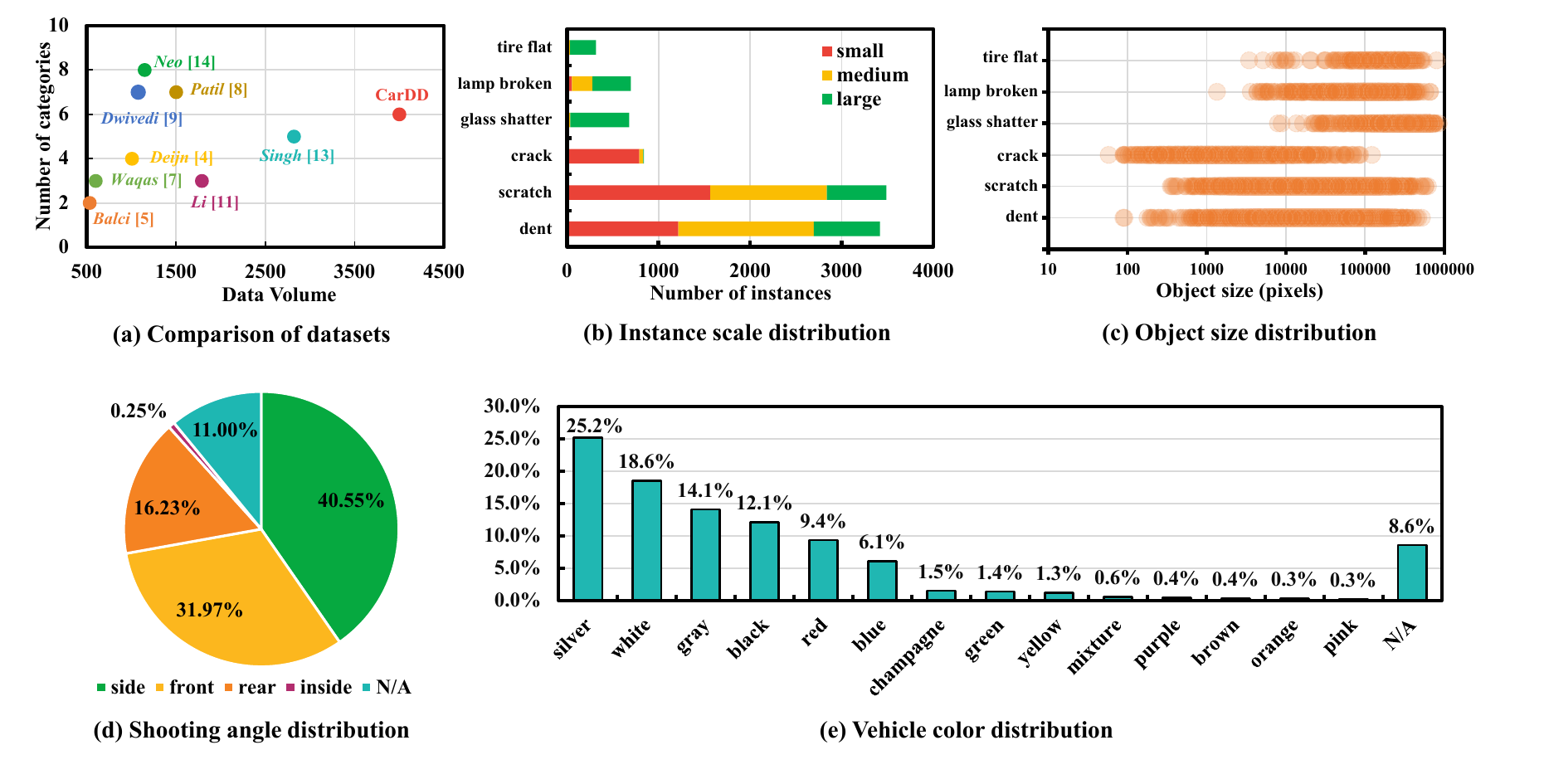}
   \caption{Dataset statistics of CarDD: (a) Comparison of datasets from the aspects of data volume and number of damage categories, (b) proportion of small, medium, and large instances of each category, (c) distribution of object size (size of the damaged area) for each class, (d) distribution of shooting angles, (e) distribution of vehicle colors. Note that the ``N/A'' refers to the proportion of pictures in which the shooting angle or the vehicle color can not be inferred.}
   \label{fig:Statistics}
   \vspace{-6mm}
\end{figure*}

\noindent {\bf Annotation Quality Control.} 
To obtain high-quality ground truths, we conduct a team containing 20 members, 5 of whom are experts in car damage assessment from the Ping An Insurance Company, and the other 15 annotators are selected from candidates as follows.
Firstly, all candidates received professional training under Ping An car insurance claim rules and the annotation guidelines as described before.
Then, the experts from the Ping An Insurance Company annotated 100 car damage images as the quality-testing set.
Each candidate took the “annotation test” by annotating this set.
Candidates who got recall scores over $90\%$ were selected to participate in the remaining annotation work.

Annotators used 27-inch display devices for annotation. The function of zooming in ($\times$ 1-8) is available. Annotating each picture takes about 3 minutes on average.

To ensure the annotation quality, experts will verify each labeled sample, and the inconsistent regions will be carefully re-annotated by experts.
Several rounds of feedback are conducted to ensure every image is well annotated.

\noindent {\bf Annotation Format.} 
The dataset contains annotations of damage categories, locations, contours, and ground-truth binary maps, as shown in Figure~\ref{fig:Pipeline}.
For object detection and instance segmentation tasks, the format of annotation files is consistent with that of the COCO dataset \cite{25}: each instance has a unique ID, along with the category of damage, point coordinates on the mask contours, and the location of the bounding box.
Tight-fitting object bounding boxes are obtained from the pixel-level polygonal segmentation masks. The bounding box is marked with the coordinates of the top left corner point and the height and width.
As for the SOD task, the format is consistent with the DUTS \cite{DUTS} dataset. The ground-truth binary maps are generated from the annotation of instance segmentation, with the category information removed. The objects in concern are activated with a grayscale value of 255, and the background is 0.

In addition, we offer abundant information about the car damage image for further possible classification tasks. An Excel file is attached to the dataset, describing the number of instances, the number of damage categories, damage severity, and the shooting angle of each image.

\subsection{Dataset Statistics}
\label{sec:dataset-statistics}

Next, we analyze the properties of the CarDD dataset. The superiorities of our dataset include the considerable dataset size, the high image quality, and the fine-grained annotations.

\noindent {\bf Data Volume.} 
The CarDD dataset contains 4,000 damaged car images with over 9,000 labeled instances covering six car damage categories. 
A comparison of existing car damage datasets from the aspects of dataset size (the number of images) and category number is shown in Figure~\ref{fig:Statistics}(a). 
Note that the specifications of these datasets are quoted directly from the corresponding reference since they are not publicly available, as mentioned in Section~\ref{sec:related-work}.
In addition, we show the distribution of shooting angles and vehicle colors in Figure~\ref{fig:Statistics}(d) and (e), respectively. 
It can be observed that CarDD covers diverse samples regarding the shooting angle and the color.

\noindent {\bf Object Size.} 
Following a similar manner in the COCO dataset, the objects in the CarDD dataset are also classified into three scales, where the ``small,'' ``medium,'' and ``large'' instances are with areas under $128^2$, over $128^2$ but under $256^2$, and over $256^2$, respectively.
As a result, the proportion of small, medium, and large instances in CarDD are $38.6\%$, $32.6\%$, and $28.8\%$, respectively.
And we report the distribution of object scale (size of the damaged area) for each class in Figure~\ref{fig:Statistics}(b) and (c).
The evaluation metrics are also designed based on this setting, see Section~\ref{sec:SD-settings}.

\noindent {\bf Image Quality.} 
High-quality images are the foundations of the annotation work.
To verify the importance of image quality to the annotation, 
we ask two groups of members to annotate images with high and low resolutions separately. The annotation results are shown in Figure~\ref{fig:Resolution}.
We can see that more instances of damage are spotted in the high-resolution images, while the annotator in the low-resolution images overlooks some tiny and blurry damages. %
This is because high-resolution images are able to exhibit more object-related legible textures and contours, facilitating the annotation work in quality and speed.
Therefore, we concentrate on collecting high-quality car damage images.

Our dataset overwhelms the GitHub dataset \cite{14} in image quality. The highest resolution of the GitHub dataset is only 414$\times$122 pixels, while the lowest resolution of our dataset is 1,000$\times$413 pixels. The average resolution of CarDD is 684,231 pixels (13.6 times higher) compared to the 50,334 pixels of the GitHub dataset. 
The average file size of CarDD is 739 KB (over $\times$82) compared to 9 KB of the GitHub dataset. 
Significantly, images in our dataset are not covered with watermarks, which ensures that damages will not be confused with watermarks.

\section{Binary Classification Experiment}
\label{sec:binary-classifier}

\subsection{Motivation} 

During data cleaning, manually picking out the candidates that can be passed to the next annotation step is extremely time-consuming since a large number of noisy samples (nearly 90\%) exist in the raw data collected from Flickr and Shutterstock.
To address this problem, we train a binary classifier to determine whether an image contains damaged cars so that we can utilize this binary classifier to help us clean the raw data automatically.

\subsection{Experimental Settings}
The training data contains 500 car damage samples that are manually selected and 500 negative samples
which are randomly picked from abandoned pictures that do not contain car damage.
And the testing data consists of another 1000 samples (500 positives + 500 negatives) non-overlapped with the training set.
We apply the VGG16 \cite{26} pre-trained on ImageNet as the backbone model and use the cross-entropy loss to fine-tune all the layers on the training set.
The model is trained with a learning rate of 0.001 and a batch size of 16 for 10 epochs.

\subsection{Results Analysis}
We use metrics of accuracy, precision, and recall to evaluate the performance of the binary classifier on the testing set.
The testing accuracy achieves $94.3\%$, and the precision and recall are $91.6\%$ and $97.6\%$, respectively.
And the false positive and false negative samples account for 4.5\% and 1.2\% of the whole testing set.
This indicates that, by the classifier, we can filter out most negative samples with an acceptable false negative rate, significantly improving the annotation efficiency.
But still, there are false positives that have similar appearances to car damages. 
Fortunately, our annotators can further check the selected positive samples (about 10\% of the raw data), resulting in a high-quality car damage dataset.

In summary, the binary classifier can largely facilitate the initial data-cleaning process since we only have to manually pick out the damaged cars from the relatively small-scale predicted positive set instead of the large-scale raw data set.

\begin{figure*}[!t]
  \centering
  \includegraphics[width=\textwidth]{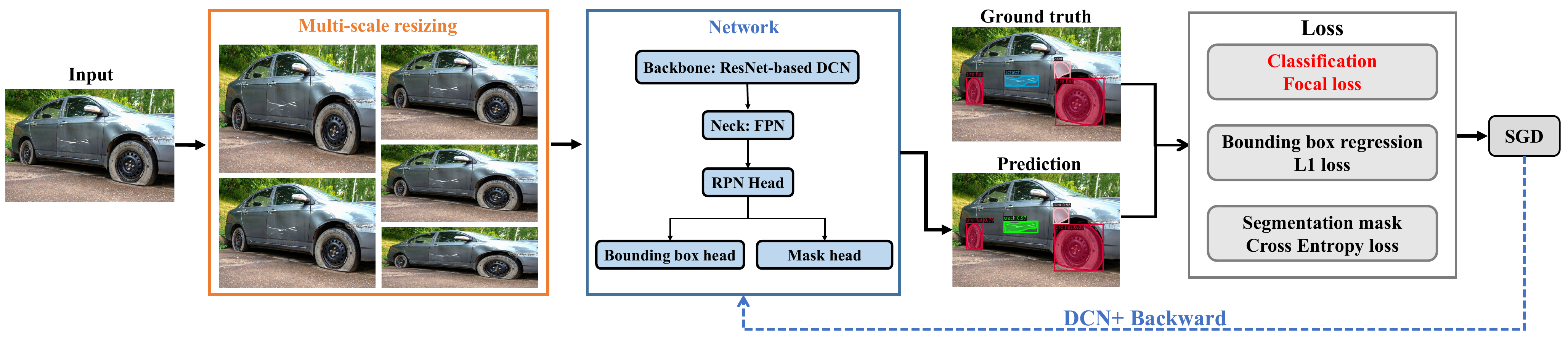}
  \caption{Flowchart of DCN+. The main improvements are highlighted, including the multi-scale resizing for data augmentation and the focal loss when calculating the classification loss. The final loss is 
  the combination of the focal loss, the $L_{1}$ loss, and the cross-entropy loss.
  }
  \label{fig:flow-chart}
\end{figure*}

\begin{table*}[!t]
\normalsize
  \caption{Ablation analysis for DCN+. The baseline is ResNet-101-based DCN. The best results in each column are highlighted in bold. AP$_{S}$, AP$_{M}$, and AP$_{L}$ stand for average precision (AP) at small, medium, and large scales, respectively. Multi-scale: multi-scale learning, glass: glass shatter, lamp: lamp broken, tire: tire flat.}
  \centering
\setlength{\tabcolsep}{6.6pt}
  \begin{tabular}{cccccccccccccccc}
    \toprule
    Baseline & Multi-scale & Focal Loss & AP & AP$_{S}$ & AP$_{M}$ & AP$_{L}$ & dent & scratch & crack & glass & lamp & tire \\
    \midrule
    \cmark &  &  & 52.5 & 19.7 & 44.8 & 66.3 & 32.0 & 24.0 & 9.8 & \textbf{92.6} & 70.4 & 86.0 \\
    \cmark & \cmark &  & 54.5 & 17.9 & 46.1 & 68.5 & 39.1 & 28.0 & 15.0 & 88.6 & \textbf{71.2} & 85.2 \\
    \cmark &  & \cmark & 56.6 & 33.2 & \textbf{47.4} & \textbf{72.2} & 39.8 & 32.6 & \textbf{16.6} & 89.6 & 71.0 & \textbf{90.1} \\
    \cmark & \cmark & \cmark & \textbf{57.0} & \textbf{34.6} & 44.0 & 71.6 & \textbf{40.5} & \textbf{34.3} & \textbf{16.6} & 89.6 & 70.8 & 90.0 \\
    \bottomrule
  \end{tabular}
  \label{tab:ablation}
\end{table*}

\begin{table}[!t]
\normalsize
  \caption{Parameters analysis of DCN+. All experiments are based on the baseline of ResNet-101 DCN. The best results in each column are highlighted in bold. AP$_{S}$, AP$_{M}$, and AP$_{L}$ stand for average precision (AP) at small, medium, and large scales, respectively.}
  \centering
  \setlength{\tabcolsep}{9.6pt}
  \begin{tabular}{cccccccccccc}
    \toprule
    $\alpha$ & $\gamma$ & AP & AP$_{S}$ & AP$_{M}$ & AP$_{L}$ \\
    \midrule
    0.25 & 1.0 & 56.2 & 23.3 & 47.1 & 61.8 \\
    0.25 & 2.0 & 56.5 & 32.1 & 47.0 & 66.6 \\
    0.25 & 3.0 & 54.8 & 24.0 & 47.3 & 60.5 \\
    0.25 & 5.0 & 48.0 & 16.3 & 38.0 & 64.8 \\
    0.50 & 1.0 & 55.6 & 20.6 & 46.5 & 69.8 \\
    0.50 & 2.0 & \textbf{56.6} & \textbf{33.2} & 47.4 & 72.2 \\
    0.50 & 3.0 & 55.3 & 27.0 & 47.4 & 60.6 \\
    0.75 & 0.5 & 55.9 & 19.5 & 47.4 & 67.3 \\
    0.75 & 1.0 & 56.3 & 28.4 & 47.3 & \textbf{74.3} \\
    0.75 & 2.0 & 56.1 & 19.7 & \textbf{50.5} & 68.3 \\
    \bottomrule
  \end{tabular}
  \label{tab:focal-loss}
\end{table}

\section{Instance Segmentation and Object Detection Experiments}
\label{sec:SD-experiments}

The objective of automatic car damage assessment is to locate and classify the damages on the vehicle and visualize them by contouring their exact locations, which naturally accords with the goals of instance segmentation and object detection. Therefore, we first experiment with the SOTA approaches designed for instance segmentation and object detection tasks. Then we introduce an improved method, called DCN+, which significantly boosts the performance on the detection of hard classes (dent, scratch, and crack). A series of ablation studies are conducted to explore the effect of each module in DCN+.

\subsection{Models}
\noindent{\bf Baselines.} 
We test five competitive and efficient models for instance segmentation and object detection, including Mask R-CNN \cite{19}, Cascade Mask R-CNN \cite{20}, GCNet \cite{21}, HTC \cite{22}, and DCN \cite{DCN} with MMDetection toolbox \cite{mmdetection}. All the initialized models have been pretrained on the COCO dataset, and we finetune them on CarDD.

\noindent{\bf DCN+.} 
In our dataset, the object scales of crack, scratch, and dent classes are diverse.
This characteristic leads the model can hardly recognize the objects of these hard classes. To solve this challenge, we introduce an improved version of DCN~\cite{DCN}, called DCN+, which can significantly increase the detection results on these hard classes. 
The flowchart of DCN+ is shown in Figure~\ref{fig:flow-chart}.
Our DCN+ involves two key techniques: multi-scale learning \cite{multiscale} that can be used to handle the scale diversity of objects and focal loss \cite{focalloss} to
enforce the model to focus on hard categories. 
Given the image, we first apply multi-scale resizing as the augmentation approach, leading to a more diverse input image size distribution without introducing additional computational or time costs.
To be specific, for multi-scale learning, we randomly resize the height of each training image in the range of [640, 1200] while keeping the width as 1333. 
Then the input will be fed into the backbone model which keeps the same as in DCN \cite{DCN}.
The network generates the predicted result, including the class, the bounding box location, and the mask of an object.
And the loss calculation is based on the predicted result and the ground truth.
The model will be optimized by minimizing the sum of the focal loss, the $L_{1}$ loss, and the cross-entropy loss.
For focal loss, we use the $\alpha$-balanced version to control the importance of different categories, formulated by
\begin{equation}
    L_{focal} = - \alpha (1 - p_{i, c})^\gamma \log(p_{i, c}),
    \label{Eq:focal-loss}
\end{equation}
where $p_{i, c}$ indicates the prediction on the ground-truth class of object $i$. 
We tried multiple combinations of $\alpha$ and $\gamma$ and chose the best one ($\alpha$=0.50 and $\gamma$=2.0) for the study of combined effect with multi-scale learning.
For training DCN+, we use an NVIDIA RTX 3090 to train the models with a batch size of 8 for 24 epochs. The learning rate is 0.01 for the first 16 epochs, 0.001 for the 17-22th epochs, and 0.0001 for the 23rd-24th epochs. Stochastic gradient descent (SGD) is adopted as the optimizer, and we set the weight decay as 0.0001 and the momentum as 0.9.

\subsection{Experimental Settings}
\label{sec:SD-settings}

\noindent {\bf Dataset Split.}
We split images in CarDD into the training set (2816 images, $70.4\%$), validation set (810 images, $20.25\%$), and test set (374 images, $9.35\%$) to make sure that instances of each category kept a ratio of 7:2:1 in the training, validation, and test sets. To avoid data leakage, we take care to minimize the chance of near-duplicate images existing across splits by explicitly removing near-duplicate (detected with the Duplicate Cleaner \cite{17}).

\noindent {\bf Evaluation Metrics.}
We report the standard COCO-style Average Precision (AP) metric which averages APs across IoU thresholds from 0.5 to 0.95 with an interval of 0.05.
Both bounding box AP (abbreviated as AP$^{bb}$) and mask AP (abbreviated as AP) are evaluated. We also report AP$_{50}$, AP$_{75}$ (AP at different IoU thresholds) and AP$_{S}$, AP$_{M}$, AP$_{L}$ (AP at different scales).
Here, the definition of AP$_S$, AP$_M$, and AP$_L$ represent AP for small objects (area\textless $128^2$), medium objects ($128^2$\textless area\textless $256^2$), and large objects (area\textgreater $256^2$), respectively, consistent with the definition in Section~\ref{sec:dataset-statistics}.
And all the results are evaluated on the \textit{test} set of CarDD.

We respectively report the mask AP of the instance segmentation branch and box AP (AP$^{bb}$) of the object detection branch in the multi-task models, see Table~\ref{tab:ap-total}. Additionally, mask AP and box AP of each category are separately reported in Table~\ref{tab:ap-category}.

\noindent {\bf Training Details.}
We use an NVIDIA Tesla P100 to train the models with a batch size of 8 for 24 epochs. The learning rate is 0.01 for the first 16 epochs, 0.001 for the 17-22th epochs, and 0.0001 for the 23rd-24th epochs. Weight decay is set to 0.0001, and momentum is set to 0.9.

\begin{table*}[!t]
\normalsize
  \caption{Comparison between our DCN+ and the state-of-the-art in terms of mask and box AP. \textbf{{\color{blue}Blue}} and \textbf{{\color{red}red}} text indicate the best result based on ResNet-50 and ResNet-101, respectively. CM-R-CNN: Cascade Mask R-CNN.}
  \centering
  \begin{tabular}{cccccccc}
    \toprule
    \textbf{Method} & \textbf{Backbone} & AP / AP$^{bb}$ & AP$_{50}$ / AP$^{bb}_{50}$ & AP$_{75}$ / AP$^{bb}_{75}$ & AP$_{S}$ / AP$^{bb}_{S}$ & AP$_{M}$ / AP$^{bb}_{M}$ & AP$_{L}$ / AP$^{bb}_{L}$ \\
    \midrule
    \multirow{1}{*}{Mask R-CNN \cite{19}} 
    & \multirow{6}{*}{ResNet-50}  & 48.4 / 49.2 & 65.5 / 66.3 & 49.7 / 52.6 & 14.3 / 16.2  & 37.6 / 41.8   & 57.3 / 55.9 \\
    {\multirow{1}{*}{CM-R-CNN \cite{20}} }
    &   & 48.6 / 50.8 & 63.5 / 64.7 & 50.3 / 53.5 & 13.1 / 16.0   & 37.1 / 36.4 & 58.7 / 58.4 \\
    \multirow{1}{*}{GCNet \cite{21}} 
    &   & 48.7 / 49.7 & 64.9 / 66.4 & 50.8 / 52.4 & 13.5 / 16.3 & 42.5 / 40.8 & 58.7 / 58.3 \\
    \multirow{1}{*}{HTC \cite{22}}   
    &   &  50.5 / 52.7 & 66.8 / 68.1 & 50.8 /  54.2  & 14.8 / 18.6 & 42.9 / 43.6 & 63.3 / 62.3   \\
    \multirow{1}{*}{DCN \cite{DCN}}   
    &   &  51.3 / 53.6 & 68.3 / 70.9 & 53.1 /  57.0  & 19.5 / 22.8 & 42.2 / 42.6 & 60.8 / 61.3   \\
    \multirow{1}{*}{DCN+ (ours)}   
    &   &  \textbf{{\color{blue}55.6}} / \textbf{{\color{blue}58.8}} & \textbf{{\color{blue}75.7}} / \textbf{{\color{blue}77.4}} & \textbf{{\color{blue}57.7}} / \textbf{{\color{blue} 63.1}}  & \textbf{{\color{blue}35.1}} / \textbf{{\color{blue}37.4}} & \textbf{{\color{blue}44.9}} / \textbf{{\color{blue}48.8}} & \textbf{{\color{blue}64.7}} / \textbf{{\color{blue}63.1}}   \\
    \hline
    \multirow{1}{*}{Mask R-CNN \cite{19}}
    & \multirow{6}{*}{ResNet-101} & 49.4 / 51.1 & 66.0 / 67.7   & 50.6 / 54.2 & 17.1 / 19.0 & 40.5 / 39.1 & 60.6 / 61.1 \\
    {\multirow{1}{*}{CM-R-CNN \cite{20}} }
    &  & 49.2 / 52.0   & 63.9 / 65.4 & 51.0 / 54.8   & 13.8 / 16.7   & 38.2 / 37.1 & 59.3 / 61.6   \\
    \multirow{1}{*}{GCNet \cite{21}} 
    &  & 50.9 / 52.6 & 67.6 / 69.3 & 52.4 / 56.6 & 18.4 / 21.2 & 42.6 / 45.8   & 58.5 / 58.4 \\
     \multirow{1}{*}{HTC \cite{22}} 
    &  &50.9  / 53.4 & 65.8 / 68.4 & 52.1 / 54.6 & 17.0 / 20.3 & 37.9 / 39.6 & 60.4 / 62.9 \\
     \multirow{1}{*}{DCN \cite{DCN}} 
    &  &52.5  / 54.3 & 68.7 / 69.8 & 54.5 / 58.3 & 19.7 / 22.7 & \textbf{{\color{red}44.8}} / 42.6 & 66.3 / 63.6 \\
     \multirow{1}{*}{DCN+ (ours)} 
    &  &\textbf{{\color{red}57.0}}  / \textbf{{\color{red}60.6}} & \textbf{{\color{red}77.7}} / \textbf{{\color{red}78.8}} & \textbf{{\color{red}58.4}} / \textbf{{\color{red}64.8}} & \textbf{{\color{red}34.6}} / \textbf{{\color{red}37.1}} & 44.0 / \textbf{{\color{red}48.0}} & \textbf{{\color{red}71.6}} / \textbf{{\color{red}66.0}} \\
    
    \bottomrule
  \end{tabular}
  \label{tab:ap-total}
\end{table*}

\begin{table*}[!t]
\normalsize
  \centering
  \caption{Comparison between our DCN+ and the state-of-the-art on each category in CarDD in terms of mask and box AP. \textbf{{\color{blue}Blue}} and \textbf{{\color{red}red}} text indicate the best result based on ResNet-50 and ResNet-101, respectively. CM-R-CNN: Cascade Mask R-CNN.}
  \begin{tabular}{cccccccc}
    \toprule
    \textbf{Method} & \textbf{Backbone} & \textbf{dent} & \textbf{scratch} & \textbf{crack} & \textbf{glass shatter} & \textbf{lamp broken} & \textbf{tire flat}  \\
    \midrule
    \multirow{1}{*}{Mask R-CNN \cite{19}}   
    & \multirow{6}{*}{ResNet-50}  & 27.0 / 27.0     &  22.2 / 25.3 & 9.3 / 19.5    & 87.8 / 87.0   & 58.5 / 53.8 & 85.8 / 82.6 \\
    \multirow{1}{*}{CM-R-CNN \cite{20}} 
    &  & 25.2 / 26.6 & 21.3 / 24.3 & 8.4 / 17.6  & 89.8 / 90.9 & 62.5 / 60.0   & 84.8 / 85.7 \\
    \multirow{1}{*}{GCNet \cite{21}} 
    &   & 26.2 / 27.2 & 21.9 / 26.0   & 9.0 / 17.0      & 89.8 / 88.5 & 61.1 / 57.0   & 84.1 / 82.2 \\
    \multirow{1}{*}{HTC \cite{22}}
    &   &   28.4   / 29.1  & 21.7 /  26.5   &  10.2   / 18.7 &  \textbf{{\color{blue}90.1}} / \textbf{{\color{blue}91.3}}  & 64.5 / 63.2 & \textbf{{\color{blue}87.9}} / \textbf{{\color{blue}87.0}}    \\
    \multirow{1}{*}{DCN \cite{DCN}}
    &   &   31.2   / 32.6  & 23.4 /  28.6   &  11.6   / 21.7 &  89.2 / 91.0  & 66.6 / 63.1 & 86.1 / 84.8    \\
    \multirow{1}{*}{DCN+ (ours)}
    &   &  \textbf{{\color{blue} 38.4}}   / \textbf{{\color{blue}39.2}}  & \textbf{{\color{blue}31.8}} /  \textbf{{\color{blue}36.6}}   & \textbf{{\color{blue} 15.4}}   / \textbf{{\color{blue}30.6}} &  89.6 / 91.2  & \textbf{{\color{blue}70.4}} / \textbf{{\color{blue}68.8}} & \textbf{{\color{blue}87.9}} / 86.4    \\
    \hline
    \multirow{1}{*}{Mask R-CNN \cite{19}} & \multirow{6}{*}{ResNet-101} & 27.9 / 29.4 & 22.3 / 25.6 &   10.3  / 20.2 & 88.9 / 88.5 & 65.4 / 62.0   & 81.6 / 80.8 \\
    \multirow{1}{*}{CM-R-CNN \cite{20}}  &  & 27.1 / 28.5 & 20.4 / 24.5 & 9.8 / 21.3  & 89.6 / 90.6 & 63.1 / 62.0   & 85.0 / 85.1   \\
    \multirow{1}{*}{GCNet \cite{21}}  &  & 29.0 / 29.6   & 22.3 / 26.5 &  10.3  /  21.6  & 89.3 / 89.3 & 66.9  / 63.7 & 87.5 / 85.0   \\
    \multirow{1}{*}{HTC \cite{22}} &  & 29.3   / 30.4  & 22.6  / 26.8  & 8.5 / 18.6  & 90.7   / \textbf{{\color{red} 92.7}}  & 65.9 / 64.3  &  88.4  /  87.5 \\ 
    \multirow{1}{*}{DCN \cite{DCN}} &  &  32.0   /  33.0  &  24.0  /  29.7  & 9.8 / 17.5  & \textbf{{\color{red} 92.6}}   / \textbf{{\color{red} 92.7}}  &  70.4 /  66.7  & 86.0  / 86.3 \\
    \multirow{1}{*}{DCN+ (ours)} &  & \textbf{{\color{red} 40.5}}   / \textbf{{\color{red} 42.2}}  & \textbf{{\color{red} 34.3}}  / \textbf{{\color{red} 42.3}}  & \textbf{{\color{red}16.6}} / \textbf{{\color{red}29.6}}  & 89.6   / 90.1  & \textbf{{\color{red} 70.8}} / \textbf{{\color{red} 69.5}}  & \textbf{{\color{red}90.0}}  / \textbf{{\color{red}90.2}} \\
    \bottomrule
  \end{tabular}
  \label{tab:ap-category}
\end{table*}

\begin{table*}[!t]
\normalsize
  \caption{Comparison between our DCN+ and the state-of-the-art in CarDD in terms of mask and box AP. The results are evaluated under the more challenging setting, in which 51 false positives are added to the test set. CM-R-CNN: Cascade Mask R-CNN. The best results in each column are highlighted in bold.}
  \centering
  \begin{tabular}{cccccccc}
    \toprule
    \textbf{Method} & \textbf{Backbone} & AP / AP$^{bb}$ & AP$_{50}$ / AP$^{bb}_{50}$ & AP$_{75}$ / AP$^{bb}_{75}$ & AP$_{S}$ / AP$^{bb}_{S}$ & AP$_{M}$ / AP$^{bb}_{M}$ & AP$_{L}$ / AP$^{bb}_{L}$ \\
    \midrule
    Mask R-CNN \cite{19} & \multirow{6}{*}{ResNet-101} & 48.7 / 50.4 & 65.0 / 66.6 & 49.8 / 53.5 & 17.0 / 19.0 & 40.1 / 38.5 & 59.8 / 60.3 \\
    CM-R-CNN \cite{20} &  & 48.7 / 51.6 & 63.3 / 64.8 & 50.5 / 54.3 & 13.7 / 16.7 & 37.9 / 37.0 & 58.7 / 61.1 \\
    GCNet \cite{21} &  & 49.0 / 50.8 & 65.3 / 66.9 & 50.3 / 54.6 & 18.1 / 21.1 & 39.3 / 43.3 & 56.9 / 56.5 \\
    HTC \cite{22} &  & 50.2 / 52.7 & 64.8 / 67.4 & 51.3 / 54.0 & 16.9 / 20.3 & 37.7 / 38.6 & 59.6 / 62.1 \\
    DCN \cite{DCN} &  & 51.9 / 53.7 & 67.9 / 69.0 & 53.9 / 57.6 & 19.3 / 22.6 & \textbf{43.3} / 42.4 & 65.6 / 62.9 \\
    DCN+ (ours) &  & \textbf{55.8} / \textbf{59.4} & \textbf{76.1} / \textbf{77.3} & \textbf{57.1} / \textbf{63.5} & \textbf{23.8} / \textbf{31.1} & 40.5 / \textbf{47.6} & \textbf{70.2} / \textbf{63.7} \\
    \bottomrule
  \end{tabular}
  \label{tab:ap-total-challenging}
\end{table*}

\subsection{Parameter Analysis}
There are four parameters in the proposed DCN+: the resize range and width in multi-scale learning, and 
the $\alpha$ and $\gamma$ in focal loss (Eq~\ref{Eq:focal-loss}).
Following the setting in~\cite{multiscale}, we fix resize range and width to [640, 1200] and 1333, respectively, and study the sensitivities of DCN+ to $\alpha$ and $\gamma$.
The baseline model is ResNet-101-based DCN.
We experiment with multiple combinations of $\alpha$ and $\gamma$, and report the  corresponding AP, AP$_{S}$, AP$_{M}$, and AP$_{L}$ in Table~\ref{tab:focal-loss}. 
We can observe that a large value of  $\alpha$ can benefit the performance in medium and large-scale object detection (see AP$_{M}$, AP$_{L}$). But a large value of $\alpha$ also hurt the small-scale detection (see AP$_{S}$).
Therefore, to balance the overall performance, we set $\alpha$=0.50 and $\gamma$=2.0 in the following experiments.

\subsection{Ablation Study}
There are two modules \textit{i.e.}, multi-scale learning (MSL) for diverse scale objects learning and focal loss for class difficulty balancing in the proposed DCN+ approach.
To investigate the effectiveness of these two modules,
we conduct a series of ablation studies by adding them to the baseline of ResNet-101 DCN, as shown in Table~\ref{tab:ablation}.
From Table~\ref{tab:ablation}, we make the following observations. 
First, adding focal loss can significantly improve the results of the hard classes. Specifically, the AP of the dent, scratch, and crack is increased by 7.8\%, 8.6\%, and 6.8\%, respectively. Second, by using the focal loss, the AP on small-scale objects (AP$_S$) is largely boosted, which mainly benefits from the gains on the dent, scratch, and crack classes. Third, by additionally injecting the MSL module, the performance in the hard classes is further improved. Our DCN+ immensely increases the baseline on the AP and $AP_S$ with an acceptable degradation on $AP_M$ and $AP_L$. These observations demonstrate the effectiveness of our DCN+ in improving the performance on hard classes, which is important for accurate and comprehensive car damage detection.
\begin{figure}[!t]
  \centering
  \includegraphics[width=\linewidth]{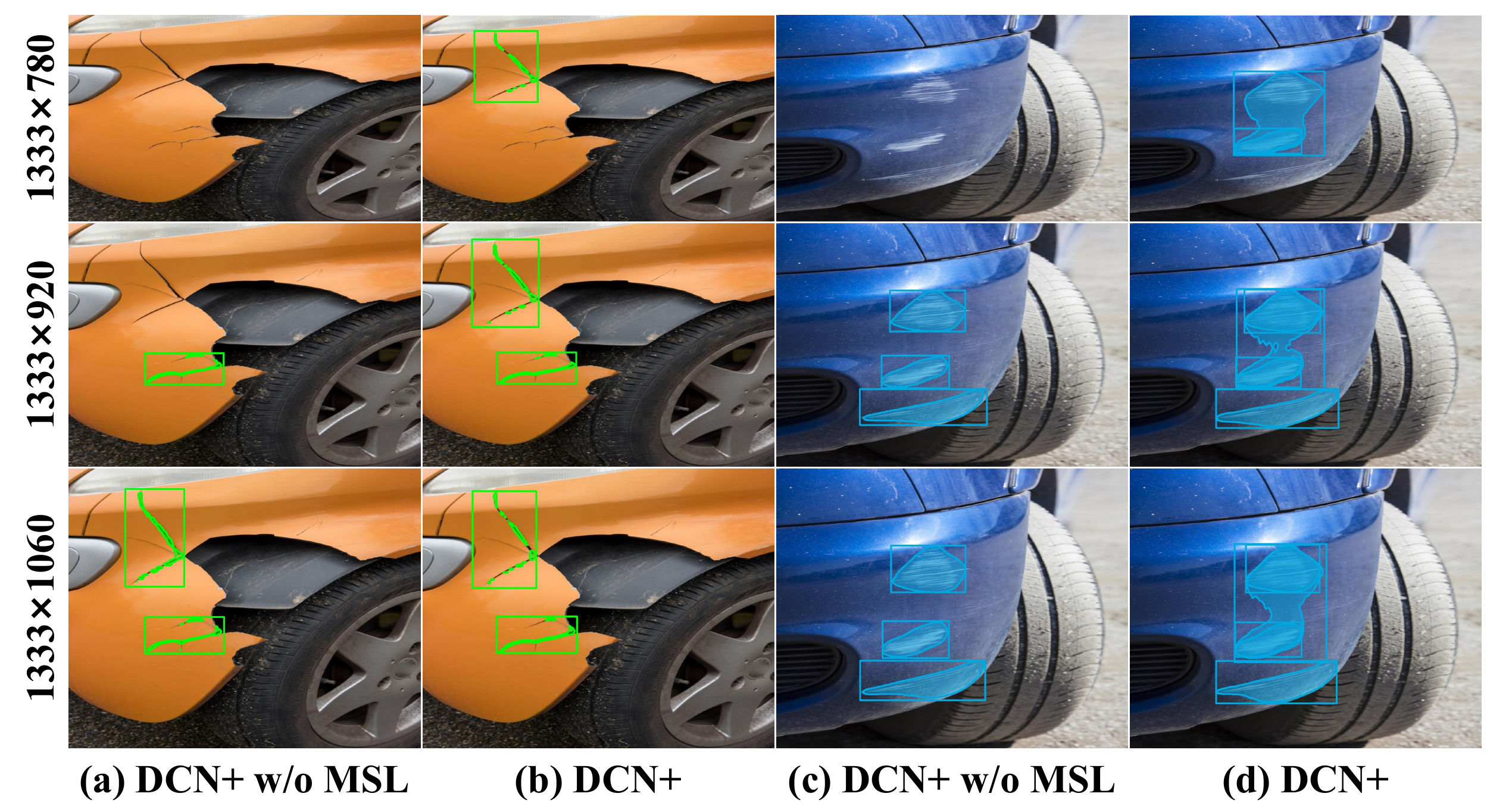}
  \caption{Visualization of the effect of multi-scale learning. (a) and (c) show the results of the model without the multi-scale learning (MSL) module, while (b) and (d) show the results with the MSL module. 
  Cracks are marked in \textcolor[rgb]{0, 1, 0}{green} and scratches are marked in \textcolor[rgb]{0, 0.66, 1}{blue}.}
  \label{fig:multiscale}
  \vspace{-6mm}
\end{figure}

In addition, to further show the effectiveness of MSL, we visualize some detection and segmentation results with and without the multi-scale learning (MSL) module based on multiple scales of inputs, as shown in Figure~\ref{fig:multiscale}.
Note that the DCN+ w/o MSL and DCN+ represent the DCN+ model without and with the MSL module.
By comparing the results of the DCN+ model under different input scales, we can see that the advantage of the MSL module is more obvious when encountering low-quality input.
For example, when the input size is 1333$\times$780, the DCN+ w/o MSL model fails to detect the damage while the model with MSL can successfully the \textit{cracks} and \textit{scratches} damage.
This indicates that the MSL module is able to address more challenging scenarios, verifying the necessity of the MSL module in the proposed DCN+ model.

\subsection{Comparison with the State-of-the-arts}
\label{sec:SD-results}

We first report the average performance on all the damage categories of different state-of-the-art methods. The results are shown in Table~\ref{tab:ap-total}. Next, we show the detailed performance of the models in each category (Table~\ref{tab:ap-category}).
These results introduce three observations. 
First, using a deeper backbone can commonly lead to higher results. 
Second, our DCN+ achieves the best results on almost all AP metrics in Table~\ref{tab:ap-total}, especially on the AP$_{S}$ and AP$^{bb}_{S}$ metrics.
Third, DCN+ significantly improves the results in the dent, scratch, and crack classes, demonstrating the effectiveness of our DCN+ in improving the performance of challenging classes.

\begin{figure*}[!t]
  \centering
   \includegraphics[width=1\textwidth, height=0.52\textwidth]{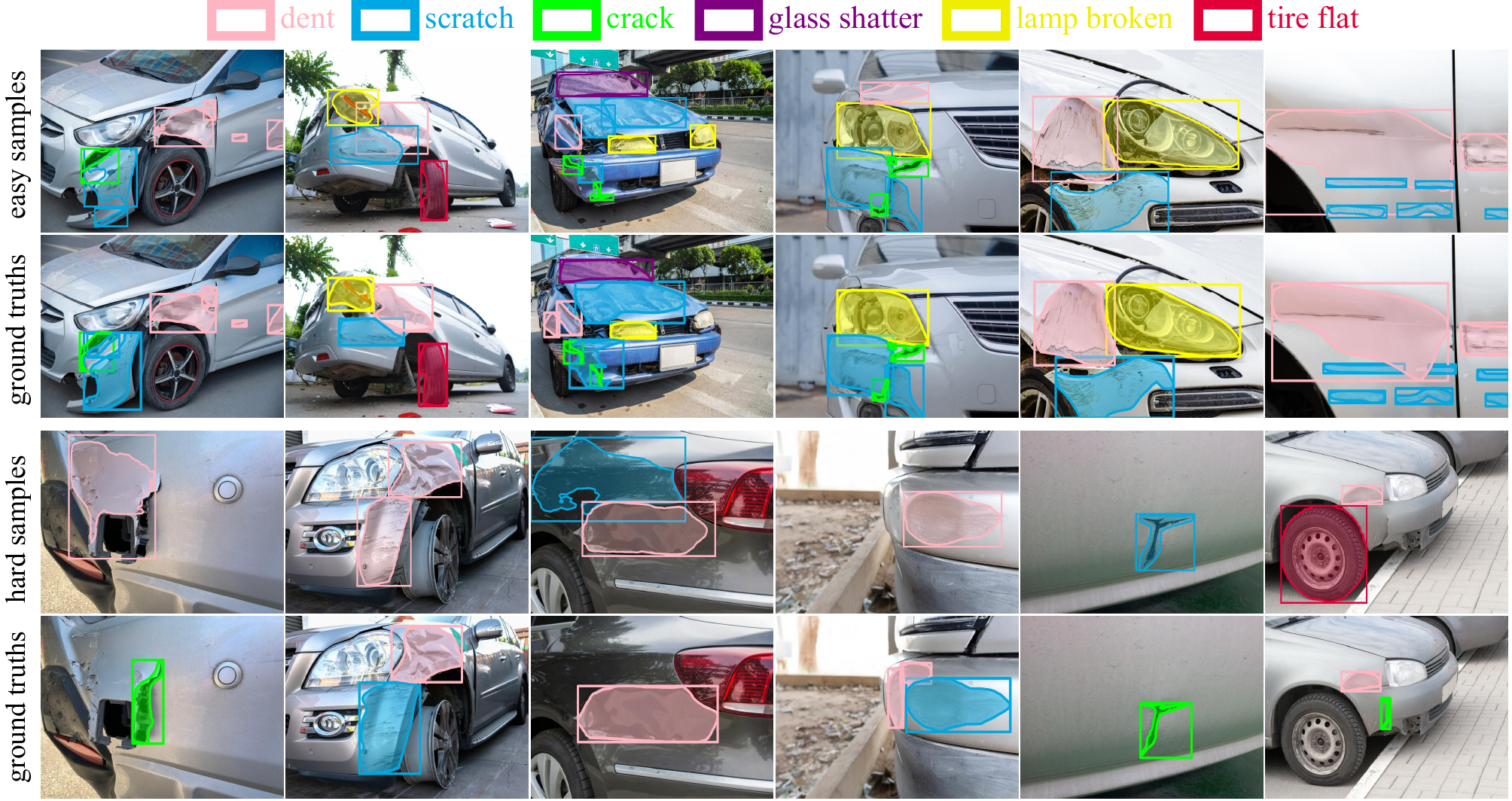}
   \caption{Samples of object detection and instance segmentation results. The first two rows are the easy samples and ground truths. The last two rows are the hard samples and ground truths.}
   \label{fig:Result-ODIS}
\end{figure*}

\subsection{Visualization}
We visualize some segmentation results from the test set (Figure~\ref{fig:Result-ODIS}).
We split the samples into two groups where the samples that can be accurately detected and segmented are called \textbf{easy samples}, and the others are called \textbf{hard samples}.
Not surprisingly, these three classes, \textit{dent, scratch,} and \textit{crack} appear in hard samples.
By carefully comparing hard and easy samples, we find that the instances in the easy samples have either a clearer boundary or a quite good shooting angle.
The bad cases include confusion between categories, mistaking metallic reflection on the car body as damages, and overlapping or mixing damages.

Through the quantitative and visualization, we can observe that the performance of these methods is not good enough, particularly on the dent, scratch, and crack, and we give the following reasons.
\textbf{First}, 
in Figure~\ref{fig:Statistics}(c), we show the object size distribution for each class.
On the one hand, it can be observed that classes \textit{dent, scratch}, and \textit{crack} have relatively various object sizes than others, meaning those classes are more diverse.
This diversity might bring new challenges to existing algorithms.
On the other hand, the shape of these three classes, \textit{i.e.}, \textit{dent, scratch,} and \textit{crack} change unreasonably.
Existing models can not carefully address this difference.
\textbf{Second}, in Figure~\ref{fig:Statistics}(b), we observe that small objects account for a large proportion in the class  \textit{dent, scratch,} and \textit{crack}, reaching over 35\%, 45\%, and 90\%, respectively,
while most of the objects in class \textit{glass shatter, tire flat} and \textit{lamp broken} are big in size.
Small objects are notoriously harder to detect than big objects. 
\textbf{Third}, the \textit{dent, scratch,} and \textit{crack} are often intertwined and look alike since they are all produced on a smooth body.
For example, both \textit{scratch} and \textit{crack} are featured with slender lines and similar colors.
The wrong prediction of one class will create a ripple effect that reaches the overlapped classes.

In summary, the diversity on both object scale and shape, small object size, and flexible boundary make \textit{dent, scratch,} and \textit{crack} more difficult to detect.

\begin{figure}[!t]
  \centering
  \includegraphics[width=0.5\textwidth, height=0.183\textwidth]{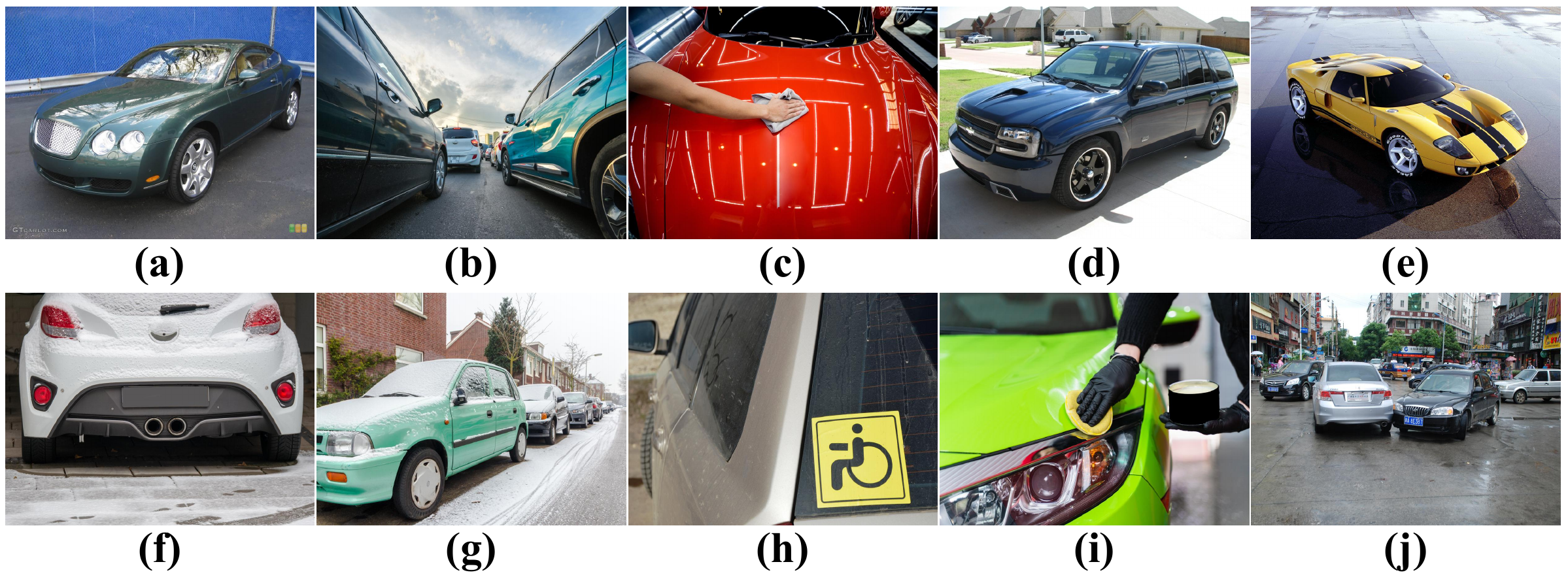}
  \caption{Some undamaged examples in the more challenging setting. (a), (b), and (c) show the metallic reflection on cars. (d) and (e) show vehicle outlines that may be confused with damages. (f), (g), and (h) show vehicles covered by snow or dirt. (i) shows a close shot. (j) shows a long shot that contains multiple cars. And the best view is in zoom.}
  \label{fig:undamaged-examples}
\vspace{-6mm}
\end{figure}

\subsection{More Challenging Setting}

In practice, a scammer may use an undamaged picture to rip off the system, which can cause economic loss for the insurance company.
To address this issue, we design a more challenging setting
where the test set contains not only the original 374 damaged samples in the original test set (see Section~\ref{sec:SD-settings}) but also another 500 undamaged samples.
Some undamaged samples are shown in Figure~\ref{fig:undamaged-examples}.
For this setting, we first use the pretrained binary classifier (in Section~\ref{sec:binary-classifier}) to detect undamaged samples from the test set. Specifically, we regard the samples with undamaged probabilities larger than 0.9 as undamaged samples. Since we use a high threshold, we can effectively avoid classifying damaged images into the undamaged category. In this manner, 449 undamaged samples are selected and discarded. Then, we send the remaining images to the car damage detection and segmentation models. We found that only 10 undamaged samples have been detected with damaged regions. 
And the comparison results are listed in Table~\ref{tab:ap-total-challenging}.
It can be seen that although the performance of all the models deteriorates in the challenging setting, the proposed DCN+ still maintains its advantages.
And compared to the setting
with only damaged images, the AP of DCN+ is only reduced by 1.2\% in this challenging setting. 
Note that, although we do not include undamaged samples during the training of car damage detection and segmentation, the
coexistence of damaged and undamaged regions within images enables the models to successfully recognize the undamaged
regions during testing. 
This experiment indicates that our
pretrained binary classifier and DCN+ model can form a double-check manner to avoid fraudulent conduct.
Additionally, we can also see that the 
$AP_{S}$ declines dramatically, which demonstrates that the model is very sensitive in detecting small-size objects in this challenging setting.

\section{Salient Object Detection Experiments}
\label{sec:SOD-experiments}

Based on the result analysis in Section~\ref{sec:SD-results}, we observe that the hard samples of the dent, scratch, and crack are considerable challenges to current methods designed for instance segmentation and object detection tasks.
The irregular shapes and flexible boundaries make it difficult to accurately detect those instances, and the similar features of the contour and color raise the possibility of confusing the classes of scratch and crack.
Aiming to deal with these challenges, we attempt to apply the salient object detection (SOD) methods for the following reasons.
First, SOD methods concentrate on refining the boundaries of salient objects, which is more suitable for segmenting objects with irregular and slender shapes.
Second, SOD methods focus on locating all the salient objects in the image without classifying the objects. For the classes with little category information (dents, scratches, and cracks), it is more important to determine the location and magnitude of the damage rather than classify them.

\subsection{Experimental Settings}
\label{sec:SOD-settings}

\noindent {\bf Dataset.}
The dataset for the SOD task is comprised of corresponding binary maps generated from the annotation of instance segmentation, as shown in Figure~\ref{fig:Visualization}. The split ratio of training, validation, and test set keeps the same as that in the detection and segmentation experiments. 

\noindent {\bf Models.}
Four SOTA methods designed for the SOD task are applied on CarDD: CSNet \cite{CSNet}, PoolNet \cite{PoolNet}, U$^{2}$-Net \cite{U2-Net}, and SGL-KRN \cite{SGL-KRN}.
We train all the models on CarDD from scratch. 
Besides, to evaluate the performance of SOD methods on each damage category, we conduct class-wise experiments, \textit{i.e.}, training and testing the models on CarDD with masks of only one class of damages activated in the binary maps.

The hyper-parameters in the training process are listed in Table~\ref{tab:hyper-parameters}.
We take adam \cite{kingma2015adam} as the optimizer for all the models.
Note that SGL-KRN and PoolNet only support a batch size of 1.
For U$^{2}$-Net, we train the model with a batch size of 12 for 140,800 iterations, equivalent to 600 epochs with 2,816 images in the training set of CarDD.

\begin{table}[!t]
  \footnotesize
  \centering
 \caption{Hyper-parameters of SOD methods. $\left[i, j\right]$ represents from the $i^{th}$ epoch to the $j^{th}$ epoch. }
 \setlength{\tabcolsep}{2pt}
  \begin{tabular}{l|c|c|c|c}
    \toprule
& SGL-KRN \cite{SGL-KRN}   & PoolNet \cite{PoolNet}   & U$^{2}$-Net \cite{U2-Net}  & CSNet \cite{CSNet} \\ \hline
Weight Decay & 5e-4      & 5e-4      & 0                             & 5e-3  \\ \hline
Batch Size   & 1         & 1         & 12                            & 16    \\  \hline
Max Epochs   & 24        & 24        & 600     & 300   \\ \hline
Learning Rate &
  \begin{tabular}[c]{@{}l@{}} $\left[1, 15\right]$: 5e-5\\ $\left[16, 24\right]$: 5e-6\end{tabular} &
  \begin{tabular}[c]{@{}l@{}} $ \left[1, 15\right]$: 5e-5\\ $\left[16, 24\right]$: 5e-6\end{tabular} &
  1e-3 &
  \begin{tabular}[c]{@{}l@{}} $\left[1, 200\right]$: 1e-4\\ $\left[201, 250\right]$: 1e-5\\ $\left[251, 300\right]$: 1e-6\end{tabular} \\
    \bottomrule
  \end{tabular}
  \label{tab:hyper-parameters}
\end{table}

\noindent {\bf Evaluation Metrics.}
To quantitatively evaluate the performance of SOD methods on CarDD, we adopt five widely used metrics, including F-measure ($F_{\beta}$) \cite{F-measure}, weighted F-measure ($F^{w}_{\beta}$) \cite{Weighted-F}, S-measure ($S_{m}$) \cite{S-measure}, E-measure ($E_{m}$) \cite{E-measure}, and mean absolute error ($MAE$).
We refer the readers to \cite{F-measure,Weighted-F,S-measure,E-measure} for more details.

\begin{figure*}[!t]
  \centering
  \hsize=\textwidth
   \includegraphics[width=1\textwidth, height=0.487\textwidth]{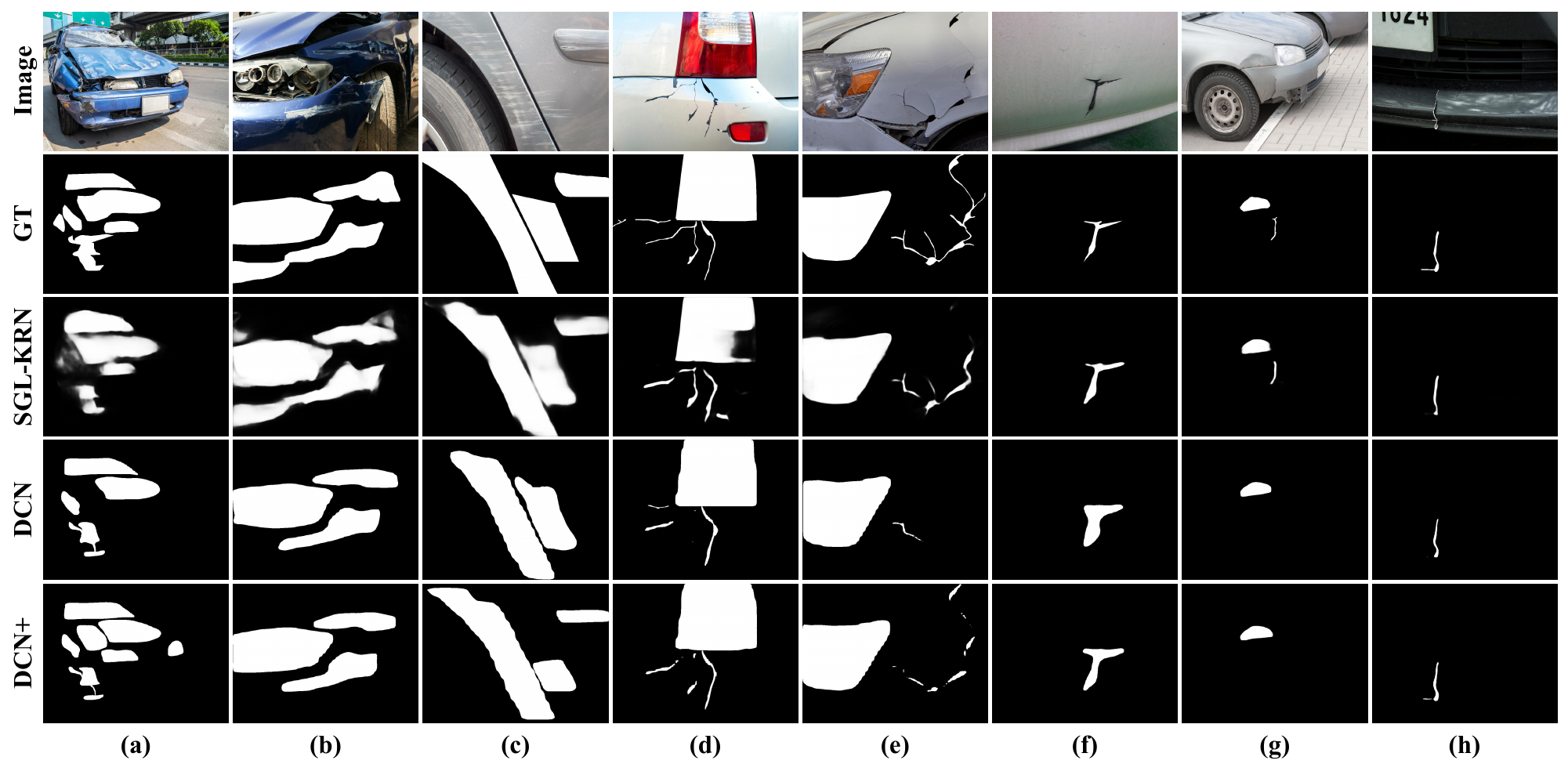}
   \caption{Visual comparison of SOD and segmentation methods. Each column shows saliency maps of one image, and each row represents one algorithm. Note that DCN \cite{DCN} and DCN+ are instance segmentation methods, and we transform their segmentation results into saliency maps for comparison.
   Columns (a), (b), and (c) contain multiple instances, columns (d) and (e) are slender objects, and columns (f), (g), and (h) are tiny objects.}
   \label{fig:Result-SOD}
\end{figure*}

\subsection{Results Analysis}
\label{sec:SOD-results}

\noindent {\bf Quantitative Comparison.} 
As shown in Table~\ref{tab:SOD-total}, we compare CSNet \cite{CSNet}, U$^{2}$-Net \cite{U2-Net}, PoolNet \cite{PoolNet}, and SGL-KRN \cite{SGL-KRN} in terms of $F_{\beta}$, $F^{w}_{\beta}$, $S_{m}$, $E_{m}$, and $MAE$. 
SGL-KRN has shown the best performance on almost all the metrics except the $S_{m}$, while PoolNet performs best on the $S_{m}$ and $MAE$.  
The benefit of PoolNet is the introduction of a global guidance module and based on PoolNet, SGL-KRN uses body-attention maps to increase the resolution of the regions related to salient objects, which results in small salient objects being magnified closer to the image size. This characteristic makes it pretty good at locating and segmenting small-size objects that account for a large proportion of the class dent, scratch, and crack in CarDD.

Furthermore, we list the 
performance of SGL-KRN on each category in Table~\ref{tab:SOD-class}.
From Table~\ref{tab:SOD-class}, we make several observations.
(a) The easy and hard classes under most metrics (especially in the first four metrics) are consistent with the findings in Table~\ref{tab:ap-category}.
(b) It is interesting that class \textit{crack} achieves the best result on the MAE metric while performance on the other four metrics is the opposite.
This is because MAE denotes the average performance between a predicted saliency map and its ground truth task, leading to it being sensitive to the object size.
In other words, the MAE metric is not suitable for measuring the model performance on car damage assessment tasks in which the object size between different categories is quite diverse.
Therefore, future works should study the SOD metrics tailored for car damage assessment tasks.
(c) Roughly speaking, it seems that the gap between the easy and hard classes is decreased in the SOD task.
To measure the gap, we calculate the standard deviation of all classes.
For instance, the standard deviation of the DCN+ model on metric mask AP is 0.28 (see Table~\ref{tab:ap-category}) while such value is reduced to 0.18 (Table~\ref{tab:SOD-class}) on $F_{\beta}$ measure.
We conjecture this is because compared to the multi-loss optimization in the detection and segmentation task, the loss in the SOD task is a binary loss simply determining whether a region is damaged or not, which maybe is much easier to be optimized for the optimizer during training.

\begin{table}[!t]
  \centering
  \caption{Quantitative salient object detection results of CSNet, U$^{2}$-Net, PoolNet, and SGL-KRN. The best results are highlighted in bold. Note that $\uparrow$ means ``the higher the better'' and $\downarrow$ means ``the lower the better.''}
  \setlength{\tabcolsep}{2mm}{
\begin{tabular}{lccccc}
\toprule
& $F_{\beta}$ $\uparrow$ & $F^{w}_{\beta}$ $\uparrow$ & $S_{m}$ $\uparrow$ & $E_{m}$ $\uparrow$ & $MAE$ $\downarrow$  \\ 
\hline
\midrule
CSNet \cite{CSNet}\tiny{TPAMI'21}       & 0.587 &	0.568 &	0.734 &	0.767 &	0.127 \\
U$^{2}$-Net \cite{U2-Net}\tiny{PR'20}   & 0.679 &	0.663 &	0.752 &	0.842 &	0.089 \\
PoolNet \cite{PoolNet}\tiny{CVPR'19}    & 0.777 &	0.733 &	\textbf{0.811} &	0.870 &	\textbf{0.071} \\
SGL-KRN \cite{SGL-KRN}\tiny{AAAI'21}    & \textbf{0.791} &	\textbf{0.744} &	0.809 &	\textbf{0.884} &	\textbf{0.071} \\
\bottomrule
\end{tabular}
}
  \label{tab:SOD-total}
\end{table}

\begin{table}[!t]
  \centering
  \caption{Quantitative salient object detection results of SGL-KRN on each class. The best results in each column are highlighted in bold. Note that $\uparrow$ means ``the higher the better'' and $\downarrow$ means ``the lower the better.''}
  \setlength{\tabcolsep}{3mm}{
\begin{tabular}{lccccc}
\toprule
& $F_{\beta}$ $\uparrow$ & $F^{w}_{\beta}$ $\uparrow$ & $S_{m}$ $\uparrow$ & $E_{m}$ $\uparrow$ & $MAE$ $\downarrow$  \\ 
\hline
\midrule
dent             & 0.698 &	0.634 &	0.776 &	0.857 &	0.060 \\
scratch          & 0.664 &	0.577 &	0.723 &	0.862 &	0.051 \\
crack            & 0.433 &	0.462 &	0.687 &	0.741 &	\textbf{0.011} \\
glass shatter    & \textbf{0.957} &	\textbf{0.952} &	0.904 &	\textbf{0.963} &	0.025 \\
lamp broken      & 0.854 &	0.838 &	0.889 &	0.930 &	0.026 \\
tire flat        & 0.936 &	0.922 &	\textbf{0.932} &	0.956 &	0.022 \\
\bottomrule
\end{tabular}
}
  \label{tab:SOD-class}
\end{table}

\noindent {\bf Visualization.}
To compare the detail-processing ability of SOD methods with instance segmentation methods, we transform the segmentation results of DCN \cite{DCN} and our DCN+ into saliency maps. 
The highlighted objects in the saliency maps have a prediction probability over 0.5.
From Figure~\ref{fig:Result-SOD}, we obtain several observations.
First, compare to DCN, the proposed DCN+ 
performs better in processing the detail of objects, especially in the hard classes like slender cracks (as shown in columns (d) and (e)).
Second,
it is obvious that SGL-KRN is powerful at distinguishing the boundaries of small and slender objects.
For instance, in columns (d),(e), and (g), SGL-KRN 
can locate more details of objects in comparison to the segmentation methods including both DCN and DCN+.
This demonstrates the superiority of the SOD methods in detail processing,
indicating that future studies can consider
investigating an appropriate way to 
combine the advantages of SOD methods and category-dependent methods to meet the requirement of realistic applications.

\section{Conclusion}
\label{sec:conclusions-and-future-work}

In this paper, we introduce CarDD, the first public large-scale car damage dataset with extensive annotations for car damage detection and segmentation. 
It contains 4,000 high-resolution car damage images with over 9,000 fine-grained instances of six damage categories and matches with four tasks: classification, object detection, instance segmentation, and salient object detection.
We evaluate various state-of-the-art models on our dataset and give a quantitative and visual analysis of the results.
We analyze difficulties in locating and classifying the categories with small and various shapes and give possible reasons.
Focusing on this challenging task, we also provide an improved 
DCN model called DCN+ to effectively improve the performance on hard classes.

CarDD will supplement existing segmentation tasks that focus on daily seen objects (such as COCO \cite{25} and Pascal VOC \cite{VOC}). 
Many research directions are made possible with CarDD, such as anomaly detection, irregular-shaped object recognition, and fine-grained segmentation.
We hope that the proposed CarDD makes a non-trivial step for the community, which can not only benefit the study of vision tasks but also facilitate daily human life.

\section*{Acknowledgments}
We appreciate Dr. Zhun Zhong for his valuable advice throughout the process of manuscript writing and revision.
We greatly thank the anonymous referees for their helpful suggestions.
Numerical computations were performed at Hefei Advanced Computing Center.

\bibliographystyle{IEEEtran}
\bibliography{IEEEexample}

\begin{thebibliography}{10}
\providecommand{\url}[1]{#1}
\csname url@samestyle\endcsname
\providecommand{\newblock}{\relax}
\providecommand{\bibinfo}[2]{#2}
\providecommand{\BIBentrySTDinterwordspacing}{\spaceskip=0pt\relax}
\providecommand{\BIBentryALTinterwordstretchfactor}{4}
\providecommand{\BIBentryALTinterwordspacing}{\spaceskip=\fontdimen2\font plus
\BIBentryALTinterwordstretchfactor\fontdimen3\font minus
  \fontdimen4\font\relax}
\providecommand{\BIBforeignlanguage}[2]{{%
\expandafter\ifx\csname l@#1\endcsname\relax
\typeout{** WARNING: IEEEtran.bst: No hyphenation pattern has been}%
\typeout{** loaded for the language `#1'. Using the pattern for}%
\typeout{** the default language instead.}%
\else
\language=\csname l@#1\endcsname
\fi
#2}}
\providecommand{\BIBdecl}{\relax}
\BIBdecl

\bibitem{01}
W.~Zhang, Y.~Cheng, X.~Guo, Q.~Guo, J.~Wang, Q.~Wang, C.~Jiang, M.~Wang, F.~Xu,
  and W.~Chu, ``Automatic car damage assessment system: Reading and
  understanding videos as professional insurance inspectors,'' in
  \emph{Proceedings of the AAAI Conference on Artificial Intelligence},
  vol.~34, no.~09, 2020, pp. 13\,646--13\,647.

\bibitem{24}
J.~Redmon, S.~Divvala, R.~Girshick, and A.~Farhadi, ``You only look once:
  Unified, real-time object detection,'' in \emph{Proceedings of the IEEE
  Conference on Computer Vision and Pattern Recognition}, 2016, pp. 779--788.

\bibitem{19}
K.~He, G.~Gkioxari, P.~Doll{\'a}r, and R.~Girshick, ``Mask r-cnn,'' in
  \emph{Proceedings of the IEEE International Conference on Computer Vision},
  2017, pp. 2961--2969.

\bibitem{02}
J.~De~Deijn, ``Automatic car damage recognition using convolutional neural
  networks,'' in \emph{2018 Internship Report MSc Business Analytics}, 2018,
  pp. 1--53.

\bibitem{03}
B.~Balci, Y.~Artan, B.~Alkan, and A.~Elihos, ``Front-view vehicle damage
  detection using roadway surveillance camera images.'' in \emph{VEHITS}, 2019,
  pp. 193--198.

\bibitem{04}
C.~Sruthy, S.~Kunjumon, and R.~Nandakumar, ``Car damage identification and
  categorization using various transfer learning models,'' in \emph{2021 5th
  International Conference on Trends in Electronics and Informatics
  (ICOEI)}.\hskip 1em plus 0.5em minus 0.4em\relax IEEE, 2021, pp. 1097--1101.

\bibitem{05}
U.~Waqas, N.~Akram, S.~Kim, D.~Lee, and J.~Jeon, ``Vehicle damage
  classification and fraudulent image detection including moir{\'e} effect
  using deep learning,'' in \emph{2020 IEEE Canadian Conference on Electrical
  and Computer Engineering (CCECE)}.\hskip 1em plus 0.5em minus 0.4em\relax
  IEEE, 2020, pp. 1--5.

\bibitem{06}
K.~Patil, M.~Kulkarni, A.~Sriraman, and S.~Karande, ``Deep learning based car
  damage classification,'' in \emph{2017 16th IEEE International Conference on
  Machine Learning and Applications (ICMLA)}.\hskip 1em plus 0.5em minus
  0.4em\relax IEEE, 2017, pp. 50--54.

\bibitem{07}
M.~Dwivedi, H.~S. Malik, S.~Omkar, E.~B. Monis, B.~Khanna, S.~R. Samal,
  A.~Tiwari, and A.~Rathi, ``Deep learning-based car damage classification and
  detection,'' in \emph{Advances in Artificial Intelligence and Data
  Engineering}.\hskip 1em plus 0.5em minus 0.4em\relax Springer, 2021, pp.
  207--221.

\bibitem{08}
N.~Patel, S.~Shinde, and F.~Poly, ``Automated damage detection in operational
  vehicles using mask r-cnn,'' in \emph{Advanced Computing Technologies and
  Applications}.\hskip 1em plus 0.5em minus 0.4em\relax Springer, 2020, pp.
  563--571.

\bibitem{09}
P.~Li, B.~Shen, and W.~Dong, ``An anti-fraud system for car insurance claim
  based on visual evidence,'' \emph{arXiv:1804.11207}, 2018.

\bibitem{10}
N.~Dhieb, H.~Ghazzai, H.~Besbes, and Y.~Massoud, ``A very deep transfer
  learning model for vehicle damage detection and localization,'' in \emph{2019
  31st International Conference on Microelectronics (ICM)}.\hskip 1em plus
  0.5em minus 0.4em\relax IEEE, 2019, pp. 158--161.

\bibitem{11}
R.~Singh, M.~P. Ayyar, T.~V.~S. Pavan, S.~Gosain, and R.~R. Shah, ``Automating
  car insurance claims using deep learning techniques,'' in \emph{2019 IEEE
  Fifth International Conference on Multimedia Big Data (BigMM)}.\hskip 1em
  plus 0.5em minus 0.4em\relax IEEE, 2019, pp. 199--207.

\bibitem{14}
``Car damage detective,'' \url{https://github.com/neokt/car-damage-detective}.

\bibitem{fine-grained}
Y.~Xiang, Y.~Fu, and H.~Huang, ``Global topology constraint network for
  fine-grained vehicle recognition,'' \emph{IEEE Transactions on Intelligent
  Transportation Systems}, vol.~21, no.~7, pp. 2918--2929, 2019.

\bibitem{vehicle-instance-retrieval}
X.~Zhang, R.~Zhang, J.~Cao, D.~Gong, M.~You, and C.~Shen, ``Part-guided
  attention learning for vehicle instance retrieval,'' \emph{IEEE Transactions
  on Intelligent Transportation Systems}, pp. 1--12, 2020.

\bibitem{vehicle-type}
Y.~Huang, Z.~Liu, M.~Jiang, X.~Yu, and X.~Ding, ``Cost-effective vehicle type
  recognition in surveillance images with deep active learning and web data,''
  \emph{IEEE Transactions on Intelligent Transportation Systems}, vol.~21,
  no.~1, pp. 79--86, 2019.

\bibitem{taillight}
H.-J. Jeon, V.~D. Nguyen, T.~T. Duong, and J.~W. Jeon, ``A deep learning
  framework for robust and real-time taillight detection under various road
  conditions,'' \emph{IEEE Transactions on Intelligent Transportation Systems},
  pp. 20\,061--20\,072, 2022.

\bibitem{licence-plate-1}
L.~Zhang, P.~Wang, H.~Li, Z.~Li, C.~Shen, and Y.~Zhang, ``A robust attentional
  framework for license plate recognition in the wild,'' \emph{IEEE
  Transactions on Intelligent Transportation Systems}, vol.~22, no.~11, pp.
  6967--6976, 2020.

\bibitem{licence-plate-2}
M.~S. Al-Shemarry, Y.~Li, and S.~Abdulla, ``An efficient texture descriptor for
  the detection of license plates from vehicle images in difficult
  conditions,'' \emph{IEEE Transactions on Intelligent Transportation Systems},
  vol.~21, no.~2, pp. 553--564, 2019.

\bibitem{12}
S.~Jayawardena \emph{et~al.}, ``Image based automatic vehicle damage
  detection,'' pp. 1--173, 2013.

\bibitem{13}
S.~Gontscharov, H.~Baumg{\"a}rtel, A.~Kneifel, and K.-L. Krieger, ``Algorithm
  development for minor damage identification in vehicle bodies using adaptive
  sensor data processing,'' \emph{Procedia Technology}, vol.~15, pp. 586--594,
  2014.

\bibitem{sod}
M.-M. Cheng, N.~J. Mitra, X.~Huang, P.~H. Torr, and S.-M. Hu, ``Global contrast
  based salient region detection,'' \emph{IEEE Transactions on Pattern Analysis
  and Machine Intelligence}, vol.~37, no.~3, pp. 569--582, 2014.

\bibitem{17}
``Duplicate cleaner,'' \url{https://www.duplicatecleaner.com}.

\bibitem{26}
K.~Simonyan and A.~Zisserman, ``Very deep convolutional networks for
  large-scale image recognition,'' \emph{arXiv:1409.1556}, 2014.

\bibitem{ImageNet}
J.~Deng, W.~Dong, R.~Socher, L.-J. Li, K.~Li, and L.~Fei-Fei, ``Imagenet: A
  large-scale hierarchical image database,'' in \emph{2009 IEEE Conference on
  Computer Vision and Pattern Recognition}.\hskip 1em plus 0.5em minus
  0.4em\relax Ieee, 2009, pp. 248--255.

\bibitem{25}
T.-Y. Lin, M.~Maire, S.~Belongie, J.~Hays, P.~Perona, D.~Ramanan,
  P.~Doll{\'a}r, and C.~L. Zitnick, ``Microsoft coco: Common objects in
  context,'' in \emph{European Conference on Computer Vision}.\hskip 1em plus
  0.5em minus 0.4em\relax Springer, 2014, pp. 740--755.

\bibitem{LFW}
G.~B. Huang, M.~Mattar, T.~Berg, and E.~Learned-Miller, ``Labeled faces in the
  wild: A database forstudying face recognition in unconstrained
  environments,'' in \emph{Workshop on Faces in 'Real-Life' Images: Detection,
  Alignment, and Recognition}, 2008, pp. 1--14.

\bibitem{DUTS}
L.~Wang, H.~Lu, Y.~Wang, M.~Feng, D.~Wang, B.~Yin, and X.~Ruan, ``Learning to
  detect salient objects with image-level supervision,'' in \emph{Proceedings
  of the IEEE Conference on Computer Vision and Pattern Recognition}, 2017, pp.
  136--145.

\bibitem{20}
Z.~Cai and N.~Vasconcelos, ``Cascade r-cnn: high quality object detection and
  instance segmentation,'' \emph{IEEE Transactions on Pattern Analysis and
  Machine Intelligence}, vol.~43, no.~5, pp. 1483--1498, 2019.

\bibitem{21}
Y.~Cao, J.~Xu, S.~Lin, F.~Wei, and H.~Hu, ``Gcnet: Non-local networks meet
  squeeze-excitation networks and beyond,'' in \emph{Proceedings of the
  IEEE/CVF International Conference on Computer Vision Workshops}, 2019, pp.
  1--10.

\bibitem{22}
K.~Chen, J.~Pang, J.~Wang, Y.~Xiong, X.~Li, S.~Sun, W.~Feng, Z.~Liu, J.~Shi,
  W.~Ouyang \emph{et~al.}, ``Hybrid task cascade for instance segmentation,''
  in \emph{Proceedings of the IEEE/CVF Conference on Computer Vision and
  Pattern Recognition}, 2019, pp. 4974--4983.

\bibitem{DCN}
J.~Dai, H.~Qi, Y.~Xiong, Y.~Li, G.~Zhang, H.~Hu, and Y.~Wei, ``Deformable
  convolutional networks,'' in \emph{Proceedings of the IEEE International
  Conference on Computer Vision}, 2017, pp. 764--773.

\bibitem{mmdetection}
K.~Chen, J.~Wang, J.~Pang, Y.~Cao, Y.~Xiong, X.~Li, S.~Sun, W.~Feng, Z.~Liu,
  J.~Xu \emph{et~al.}, ``Mmdetection: Open mmlab detection toolbox and
  benchmark,'' \emph{arXiv:1906.07155}, 2019.

\bibitem{multiscale}
B.~Singh, M.~Najibi, and L.~S. Davis, ``Sniper: Efficient multi-scale
  training,'' \emph{Advances in Neural Information Processing Systems},
  vol.~31, pp. 1--11, 2018.

\bibitem{focalloss}
T.-Y. Lin, P.~Goyal, R.~Girshick, K.~He, and P.~Doll{\'a}r, ``Focal loss for
  dense object detection,'' in \emph{Proceedings of the IEEE International
  Conference on Computer Vision}, 2017, pp. 2980--2988.

\bibitem{CSNet}
M.-M. Cheng, S.-H. Gao, A.~Borji, Y.-Q. Tan, Z.~Lin, and M.~Wang, ``A highly
  efficient model to study the semantics of salient object detection,''
  \emph{IEEE Transactions on Pattern Analysis and Machine Intelligence},
  vol.~44, no.~11, pp. 8006--8021, 2021.

\bibitem{PoolNet}
J.-J. Liu, Q.~Hou, M.-M. Cheng, J.~Feng, and J.~Jiang, ``A simple pooling-based
  design for real-time salient object detection,'' in \emph{Proceedings of the
  IEEE/CVF Conference on Computer Vision and Pattern Recognition}, 2019, pp.
  3917--3926.

\bibitem{U2-Net}
X.~Qin, Z.~Zhang, C.~Huang, M.~Dehghan, O.~R. Zaiane, and M.~Jagersand,
  ``U2-net: Going deeper with nested u-structure for salient object
  detection,'' \emph{Pattern Recognition}, vol. 106, pp. 1--12, 2020.

\bibitem{SGL-KRN}
B.~Xu, H.~Liang, R.~Liang, and P.~Chen, ``Locate globally, segment locally: A
  progressive architecture with knowledge review network for salient object
  detection,'' in \emph{Proceedings of the AAAI Conference on Artificial
  Intelligence}, vol.~35, no.~4, 2021, pp. 3004--3012.

\bibitem{kingma2015adam}
D.~P. Kingma and J.~Ba, ``Adam: A method for stochastic optimization,'' in
  \emph{ICLR (Poster)}, 2015, pp. 1--15.

\bibitem{F-measure}
R.~Achanta, S.~Hemami, F.~Estrada, and S.~Susstrunk, ``Frequency-tuned salient
  region detection,'' in \emph{2009 IEEE Conference on Computer Vision and
  Pattern Recognition}.\hskip 1em plus 0.5em minus 0.4em\relax IEEE, 2009, pp.
  1597--1604.

\bibitem{Weighted-F}
R.~Margolin, L.~Zelnik-Manor, and A.~Tal, ``How to evaluate foreground maps?''
  in \emph{Proceedings of the IEEE Conference on Computer Vision and Pattern
  Recognition}, 2014, pp. 248--255.

\bibitem{S-measure}
D.-P. Fan, M.-M. Cheng, Y.~Liu, T.~Li, and A.~Borji, ``Structure-measure: A new
  way to evaluate foreground maps,'' in \emph{Proceedings of the IEEE
  International Conference on Computer Vision}, 2017, pp. 4548--4557.

\bibitem{E-measure}
D.-P. Fan, C.~Gong, Y.~Cao, B.~Ren, M.-M. Cheng, and A.~Borji,
  ``Enhanced-alignment measure for binary foreground map evaluation,'' in
  \emph{IJCAI}, 2018, pp. 698--704.

\bibitem{VOC}
M.~Everingham, L.~Van~Gool, C.~K. Williams, J.~Winn, and A.~Zisserman, ``The
  pascal visual object classes (voc) challenge,'' \emph{International Journal
  of Computer Vision}, vol.~88, no.~2, pp. 303--338, 2010.

\end{thebibliography}

\end{document}